\documentclass[fleqn,10pt]{wlscirep}
\usepackage[utf8]{inputenc}
\usepackage[T1]{fontenc}
\usepackage{amsmath}
\usepackage{subcaption}
\usepackage{graphicx}
\usepackage[numbers]{natbib}
\title{Hidden Dynamics of Massive Activations in Transformer Training}
\usepackage{float}
\usepackage{booktabs}
\usepackage{hyperref}
\hypersetup{
    colorlinks=true,
    linkcolor=blue,
    filecolor=magenta,      
    urlcolor=cyan,
    pdftitle={Overleaf Example},
    pdfpagemode=FullScreen,
    }
    
\author[1, +]{Jorge Gallego-Feliciano}
\author[1, +]{S. Aaron McClendon}
\author[1]{Juan Morinelli}
\author[2]{Stavros Zervoudakis}
\author[2, *]{Antonios Saravanos}

\affil[1]{Aimpoint Digital Labs, Atlanta, GA, USA}
\affil[2]{New York University, New York, NY, USA}

\affil[+]{these authors contributed equally to this work}
\affil[*]{please direct correspondence to: Dr. Antonios Saravanos (saravanos@nyu.edu)}


\begin{abstract}
We present the first comprehensive analysis of massive activation development throughout transformer training, using the Pythia model family as our testbed, and release our full dataset publicly to support further research. Through systematic analysis of various model sizes across multiple training checkpoints, we demonstrate that massive activation emergence follows highly predictable mathematical patterns that can be accurately modeled using an exponentially-modulated logarithmic function with five key parameters. Additionally, We develop a machine learning framework to predict these mathematical parameters from architectural specifications alone, achieving high accuracy for steady-state behavior and moderate accuracy for emergence timing and magnitude. These findings enable architects to predict and potentially control key aspects of massive activation emergence through design choices, with significant implications for model stability, training cycle length, interpretability, and optimization. Our findings demonstrate that the emergence of massive activations is governed by model design and can be anticipated, and potentially controlled, before training begins. Code is available at \url{https://github.com/Aimpoint-Digital/massive-activations-fork}
\end{abstract}
\begin{document}

\flushbottom
\maketitle

\thispagestyle{empty}

\begin{figure}[h]

  \centering
  \begin{subfigure}[c]{0.3\textwidth}
    \centering
    \includegraphics[width=\linewidth]{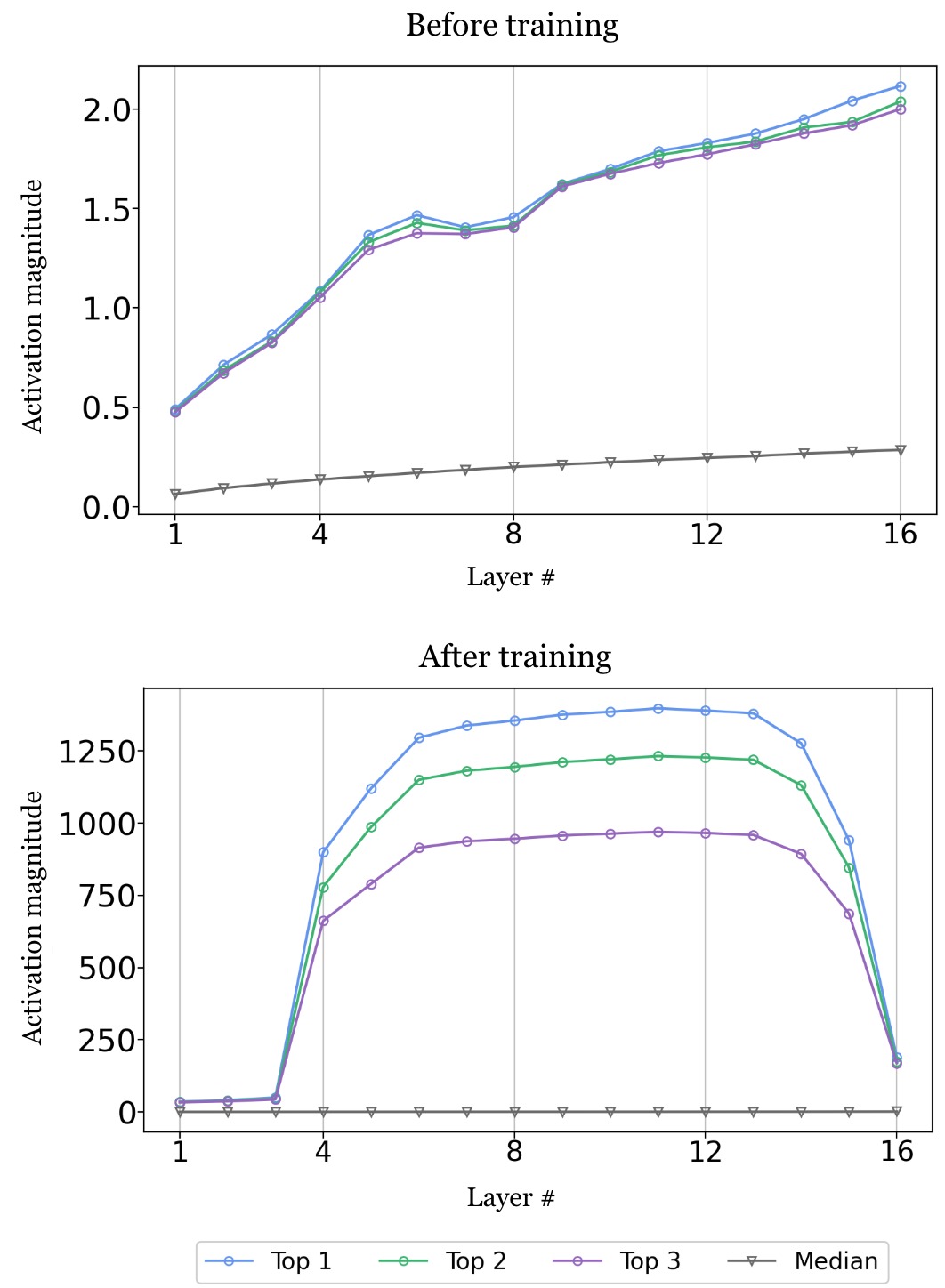}
    \caption{}
    \label{fig:r2-vertical-plot}
  \end{subfigure}
  \begin{subfigure}[c]{0.3\textwidth}
    \centering
    \includegraphics[width=\linewidth]{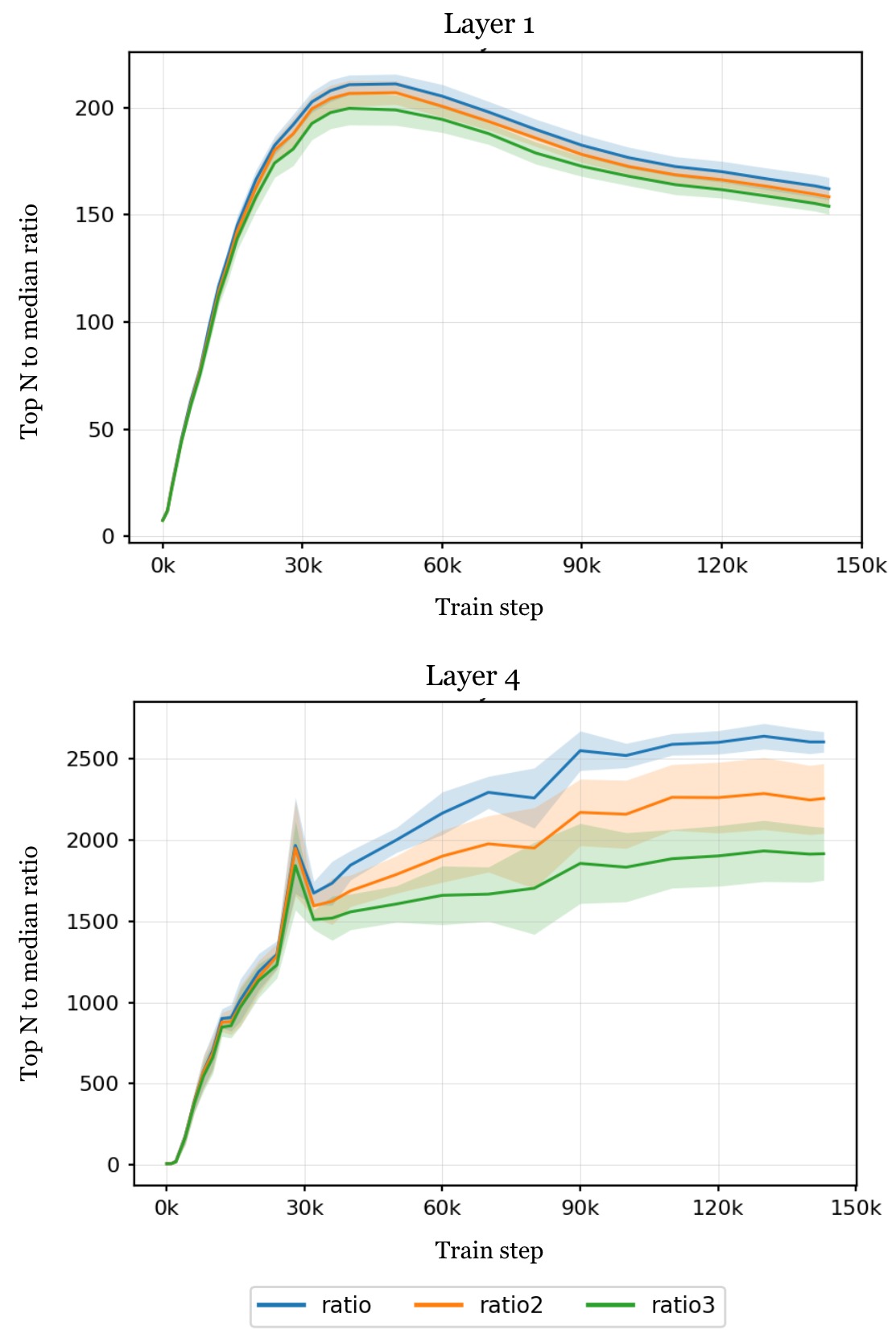}
    \caption{}
    \label{fig:r2-heatmap}
  \end{subfigure}
  \begin{subfigure}[c]{0.3\textwidth}
    \centering
    \includegraphics[width=\linewidth]{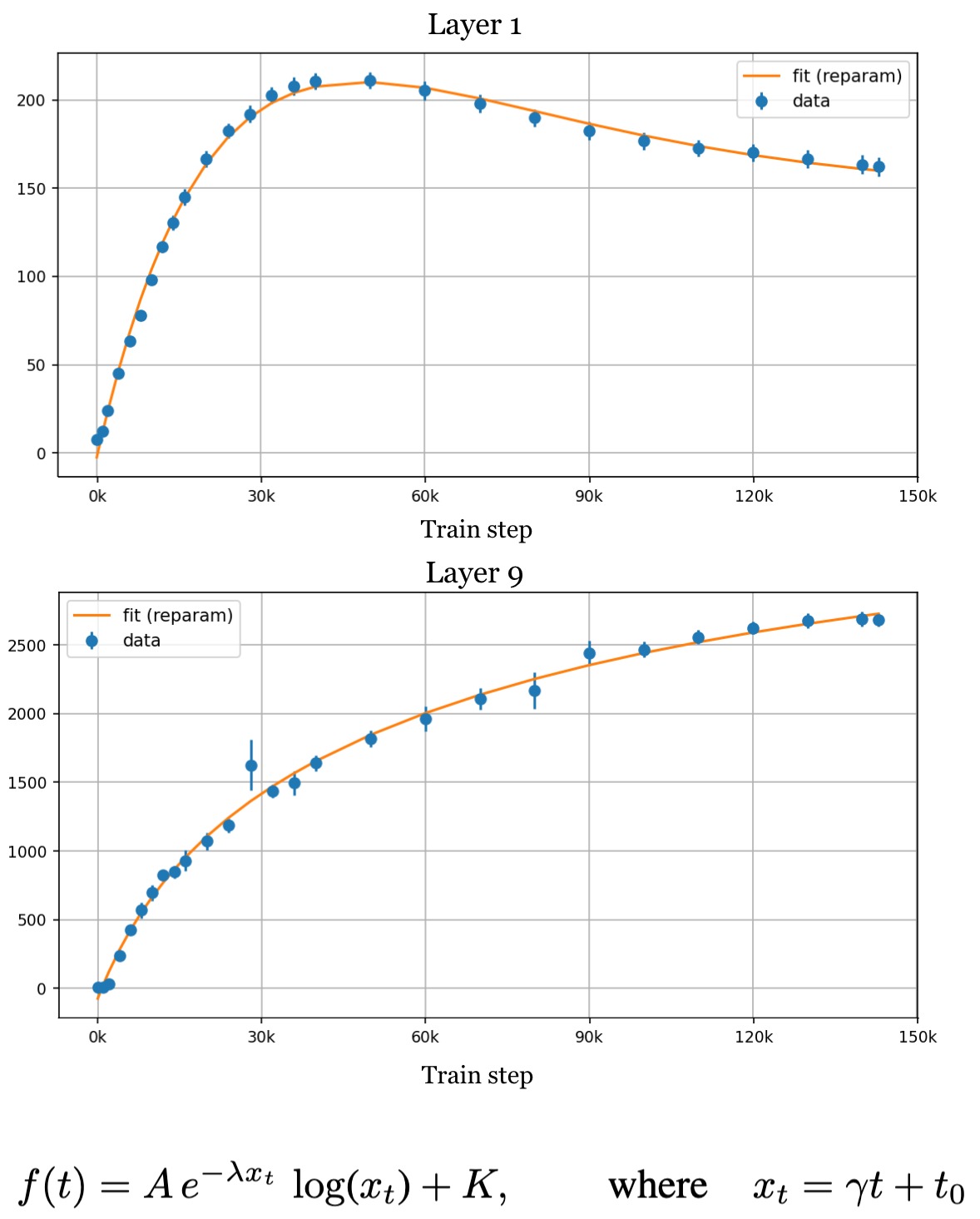}
    \caption{}
    \label{fig:r2-heatmap}
  \end{subfigure}
  \caption{Example analysis of massive activation development in Pythia 1B. \textbf{a)} Top 3 activations and median for each layer before and after training, showing that MAs develop during training. \textbf{b)} Evolution of the top three to median activation ratios during training for two example layers. \textbf{c)} 5-parameter model fits the evolution of MA with an $R^2 > 0.99$ over all layers -- here shown 2 fits.}
\end{figure}

\section*{Introduction}
Transformers have become the dominant architecture for large-scale language models, particularly through decoder-only designs for generative tasks \cite{brown2020language, biderman2023pythiasuiteanalyzinglarge}. A decoder-only transformer comprises a stack of layers that update a $d$-dimensional hidden state (the ``residual stream'') via self-attention and feed-forward sublayers, coupled with residual connections and normalization for stable information flow \cite{zhang2023dive}. Self-attention contextualizes token representations, while feed-forward networks apply position-wise transformations that build the high-dimensional features used for prediction.

A notable phenomenon in these models is the emergence of \emph{massive activations}: individual neuron activations that exceed typical magnitudes within a layer by factors of $10^3$--$10^4$ \cite{vaswani2017attention, Sun:massiveactivations:2024}. Whereas activations are generally content-dependent, these extreme values can remain nearly constant across inputs and act as implicit bias terms that steer attention toward particular tokens. Massive activations have practical consequences for quantization \cite{nrusimha2024activation_regularization,dettmers2022llmint8}, inference optimization \cite{dettmers2022llmint8,ma2024activationsmattertrainingfreemethods,szatkowski_exploiting_2024, yu2025superweightlargelanguage}, and training stability \cite{narang2024deepseek_v3}. Prior work shows that removing them can cause model failure, while replacing them with mean values can preserve functionality \cite{Sun:massiveactivations:2024}, and selectively amplifying high-impact activations can promote Chain-of-Thought reasoning without reinforcement learning \cite{zhao2025activation}. Recent analyses further trace their roots to specific architectural components \cite{an2025systematic,he2024understanding}, but key questions about their emergence during training remain open.

Existing mitigation strategies largely operate after massive activations appear. Proposed approaches include modifying attention nonlinearities (e.g., Softpick, softmax-1, clipped softmax, gated attention) \cite{zuhri2025softpick, enigmallm, bondarenko2023quantizable}, training-time interventions such as MacDrop \cite{oh2025housecardsmassiveweights}, and explicit outlier handling methods like DuQuant \cite{lin2024duquant}. Normalization placement has also been implicated: schemes such as Peri-LayerNorm, which normalize both before and after sublayers, can stabilize activation variance and gradients relative to standard Pre-LayerNorm designs \cite{kim2025peri}. Broader evidence suggests that not all outliers are harmful and that combining strategies can suppress damaging extremes while preserving downstream performance \cite{owen2025refinedanalysismassiveactivations}. Related phenomena appear in Vision Transformers, where high-norm ``artifact'' tokens can distort attention; introducing dedicated ``register'' tokens can absorb these activations and improve feature and attention maps \cite{darcet2023vision, gan2025unleashing}.

Despite these advances, most existing interventions remain fundamentally reactive, addressing the symptoms of massive activations only after they have emerged. Comparatively little is known about the \emph{temporal} dynamics of massive activations: when they first appear, how they evolve across layers and training stages, and whether their properties can be anticipated from architectural choices. A mechanistic account of their development would clarify representation formation \cite{xu2024tracking}, information propagation, and implicit biases, while also informing training diagnostics and architectures optimized for quantization and efficiency \cite{jin2025massivevalues, ma2024activationsmattertrainingfreemethods, yue2024wkvquant, yang2024mitigating}.

To address these gaps, this work studies massive activation development across training in the EleutherAI Pythia suite \cite{biderman2023pythiasuiteanalyzinglarge}, comprising 9 decoder-only transformers from 14M to 12B parameters with over 150 checkpoints per model. This setting enables a controlled analysis of when and how massive activations arise, how their trajectories depend on scale, and how well they can be predicted from architectural parameters such as depth, hidden size, and attention head count. We then provide a quantitative characterization of these trajectories, introduce a unified mathematical framework for modeling massive activation emergence, and discuss implications for transformer design, interpretability, and optimization.

\section*{Preliminary}
This section establishes the mathematical framework and key definitions necessary for analyzing massive activation dynamics during transformer training. We focus on decoder-only transformer architectures, as exemplified by the Pythia model family studied in this work.

\subsection*{Transformer architecture and hidden states}

We consider decoder-only transformer models composed of $L$ residual blocks. Each layer $\ell \in L$ receives a hidden state $h_{\ell-1} \in \mathbb{R}^{S \times d}$ and produces an updated hidden state:
\begin{equation}
h_\ell = h_{\ell-1} + \mathcal{F}_\ell(h_{\ell-1})
\end{equation}

\noindent where $\mathcal{F}_\ell$ includes both multi-head self-attention and MLP submodules. Throughout this paper, we denote by $h_\ell$ the post-residual hidden state, i.e., the output after the residual summation. As in \cite{Sun:massiveactivations:2024}, we do not consider intermediate computations within $\mathcal{F}_\ell$ unless explicitly stated.

An \textit{activation} refers to a specific scalar element of a hidden state tensor $h_\ell$. For a model processing a sequence of $S$ tokens with $d$ hidden dimensions, each layer's output $h_\ell \in \mathbb{R}^{S \times d}$ contains $S \cdot d$ scalar activations. In this work, we focus exclusively on the scalar values in $h_\ell$, rather than weights, attention logits, or intermediate MLP states. 

\subsection*{Massive activations}
\label{subsec:massive_activations}

Following the definition introduced in \cite{Sun:massiveactivations:2024}, we refer to certain rare, abnormally large activations as \textit{massive activations} (MAs). They propose a loose rule of thumb, and consider a scalar activation $a \in h_\ell$ to be massive if:

\begin{equation}
|a| > 100 \quad \text{and} \quad \frac{|a|}{\mathrm{median}(|h_\ell|)} \geq 1000.
\end{equation}

These activations have been observed to occur consistently at a small set of fixed feature dimensions and are often associated with the initial token or delimiter tokens in the input sequence (e.g., ``.'' or ``\textbackslash n''). While small in number, they are disproportionately large—often exceeding the median activation by four or more orders of magnitude—and have been shown to function as implicit bias terms in the model's computation. 

The original definition in \cite{Sun:massiveactivations:2024}, which includes a hard threshold of $|a| > 100$, does not generalize well to smaller models. For instance, in Pythia-14M (Figure~\ref{fig:ma_at_start_end}), the top activation magnitudes at layer 3 clearly dominate all others in the model, exhibiting the characteristic sharp spike associated with massive activations. Yet their absolute values remain well below 100, and their ratio to the median is also far less than $10^3$.

Despite this, the behavioral pattern is qualitatively similar to that observed in larger models: a small number of activations attain disproportionately high values, persist across tokens and inputs, and concentrate in specific feature dimensions. To better capture this effect across model scales, we relax the definition and focus instead on the top largest activation, which in the case of smaller models, even without reaching the thresholds from the previous definition, exhibit the same patterns as those activations in larger models.

\begin{figure}[h]
    \centering
    \includegraphics[width=0.7\linewidth]{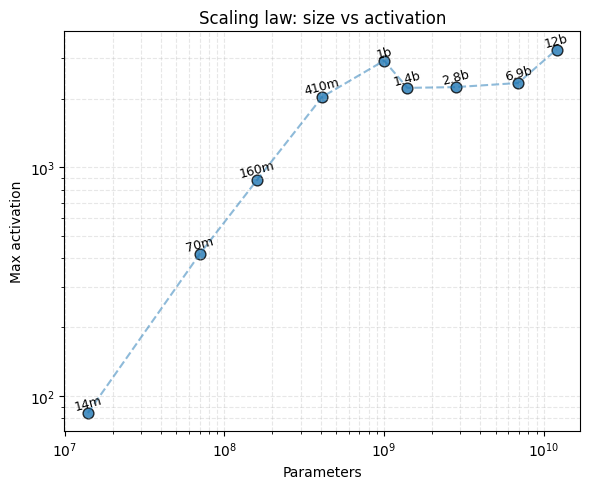}
    \caption{Plot of transformer parameter count vs value of the top activation to median ratio per model, in each respective final model checkpoint.}
    \label{fig:scaling}
\end{figure}

Figure~\ref{fig:scaling} illustrates the scaling relationship between model size and the maximum observed activation magnitude (averaged across samples in $X$). We observe a steep rise in activation magnitudes from 14M to 1B parameters, followed by a plateau and a secondary increase beyond 6B. This suggests that massive activations emerge gradually with scale and stabilize in prevalence or intensity past a certain model size. 

We study a transformer, along with its training checkpoints, denoted as $M_t$, for $0 \leq t \leq T$, where each $M_t$ is a transformer model with the GPT-NeoX architecture \cite{biderman2023pythiasuiteanalyzinglarge}. The index $t$ represents the training step, ranging from $0$ to $143{,}000$. EleutherAI released 154 checkpoints at regular intervals: multiples of 1000 steps for the full training duration, with additional higher-resolution checkpoints at powers of 2 up to step 512 for detailed analysis of early training dynamics. Each model $M_t$ consists of $L$ decoder layers. Passing an input sequence $x$ through $M_t$ yields a series of hidden states:

\begin{equation}
h_\ell(M_t, x) \in \mathbb{R}^{S \times d}
\end{equation}
where $\ell \in \{1, \ldots, L\}$ is the layer index, $S$ is the sequence length, and $d$ is the hidden dimension. For brevity, we denote activations as $h_{\ell,t}(x) := h_\ell(M_t, x)$.

\subsection*{Computing massive activations during training}

Massive activations are defined based on the \textit{top} values in each layer relative to the \textit{median}. To clarify, we characterize the activations being measured as the final hidden state output from each decoder layer, which represents the post-residual activations after both the self-attention and MLP (feed-forward) components have been applied. These correspond to the $h_\ell$ values in our notation and are the inputs to the subsequent layer. Let the following denote scalar quantities:

\begin{equation}\label{eq:median}
    h^{\text{median}}_{\ell,t}(x): \text{ the median value of } |h_{\ell,t}(x)| 
\end{equation}

\begin{equation}\label{eq:top-i}
    h^{\text{max}}_{\ell,t}(x): \text{ the largest value in } |h_{\ell,t}(x)|
\end{equation}

\begin{equation}\label{eq:ratio2}
    r_{\ell,t}(x):= \frac{h^{\text{max}}_{\ell,t}(x)}
    {h^{\text{median}}_{\ell,t}(x)} : \text{ ratio of the largest activation to the median}
\end{equation}

\noindent Since activations depend on the input, we evaluate them over a distribution $\mathcal{X}$ of realistic inputs. We define $\mathcal{X}$ to contain natural language sentences representative of real-world usage, excluding out-of-distribution inputs that would yield unpredictable behavior. We define the expected activation over $\mathcal{X}$ as:
\begin{equation}
    H_{\ell,t}(\mathcal{X}) := \mathbb{E}_{x \sim \mathcal{X}} [h_{\ell,t}(x)] \quad \text{and approximate it with} \quad \tilde{h}_{\ell,t}(X) := \frac{1}{|X|} \sum_{x \in X} h_{\ell,t}(x)
    \label{eq:approximating-mas}
\end{equation}

We similarly define $\tilde{h}^{\text{median}}_{l,t}$, $\tilde{h}^{\text{max}}_{l,t}$, and $\tilde{r}_{l,t}$.

\noindent In practice, $X$ is a random sample of 10 sequences from the RedPajama dataset \cite{together2023redpajama_sample}. We measure the variance on \ref{eq:approximating-mas} corresponding to our sample size, and report on precision and confidence intervals in the results section. Our practical experiments and prior work \cite{Sun:massiveactivations:2024} support that massive activation patterns have low variance across inputs, which justifies the sample size. 

\subsection*{Mathematical modeling of massive activation evolution}

To track the development of MAs over time, we construct a time series:

\begin{equation}\label{eq:ratio}
r_\ell := \left( \tilde{r}_{\ell,t}(x) \right)_{t \in T}
= \left( 
\frac{\tilde{h}^{\text{max}}_{\ell,t}(x)}{
\tilde{h}^{\text{median}}_{\ell,t}(x)}
\right)_{t \in T}
\end{equation}

\noindent for each layer $l$. These series are smooth and exhibit consistent patterns across model sizes and layers, suggesting the potential for a generalizable predictive model. Throughout the paper, we set the variable $i$ in $r_l$ to be 1, or, equivalently, we measure the max activation value as in Equation \ref{eq:ratio2}. 


\begin{figure}[h]
    \centering
    \includegraphics[width=0.95\linewidth]{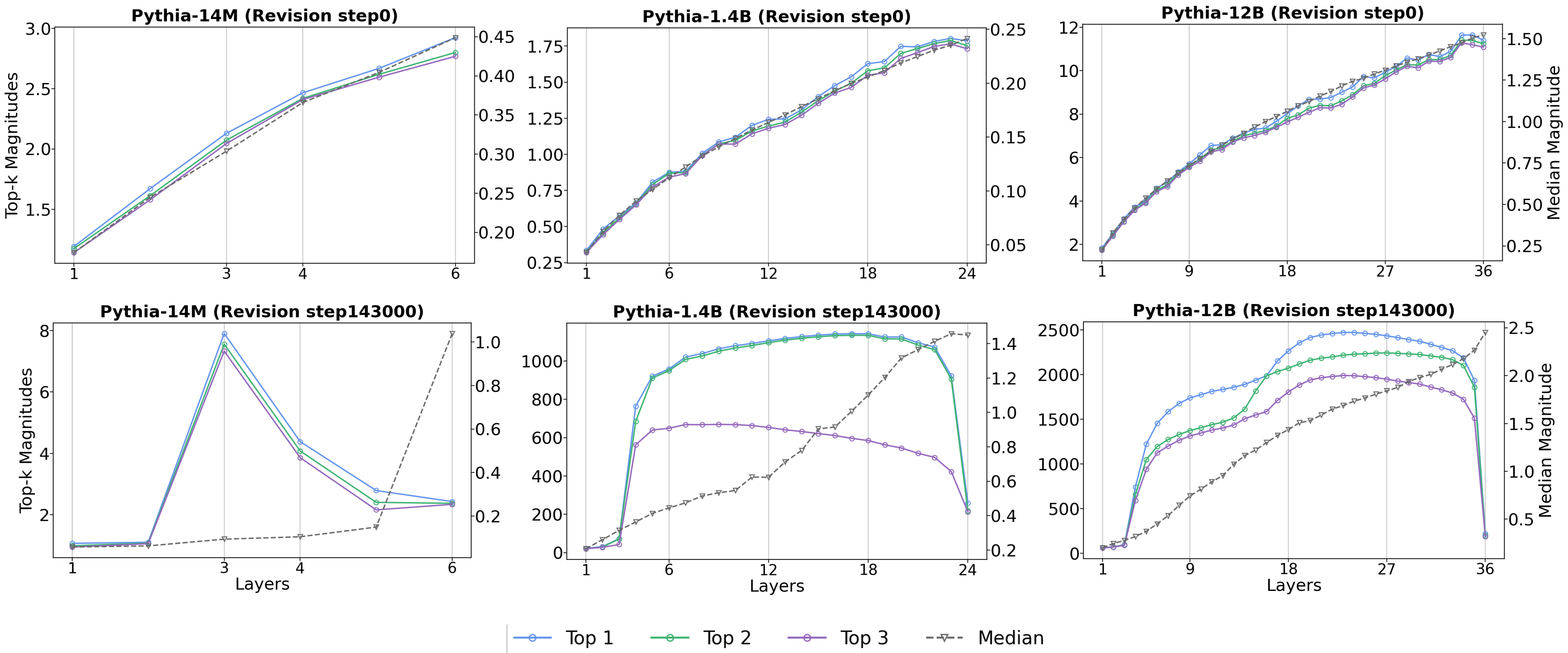}
    \caption{Top activation magnitudes per layer in models Pythia-14M, Pythia-1.4B and Pythia-12B at revision step 0 and 143000, which correspond to the start and end of training. Pythia-14M reaches a top 1 to median ratio of ~83, Pythia-1.4B reaches ~2350, and Pythia-12B reaches ~3200.}
    \label{fig:ma_at_start_end}
\end{figure}

\section*{Results}

This section reports two main findings on MAs. First, we trace how MA magnitudes rise and fall throughout training and fit these curves with an accurate predictive model. Second, we show how key architectural choices—layer depth, hidden width, and head count—shape those trajectories, revealing design-level predictors of when and how large MAs will become.

\subsection*{Evolution of massive activations during training}

We now focus on the evolution of the ratio of the top 1 activation to the median (Equation \ref{eq:ratio}), a magnitude that characterizes massive activations. For convenience, throughout this section, we refer to this quantity simply as `massive activations'. 

\begin{figure}[h]
    \centering
    \includegraphics[width=0.95\linewidth]{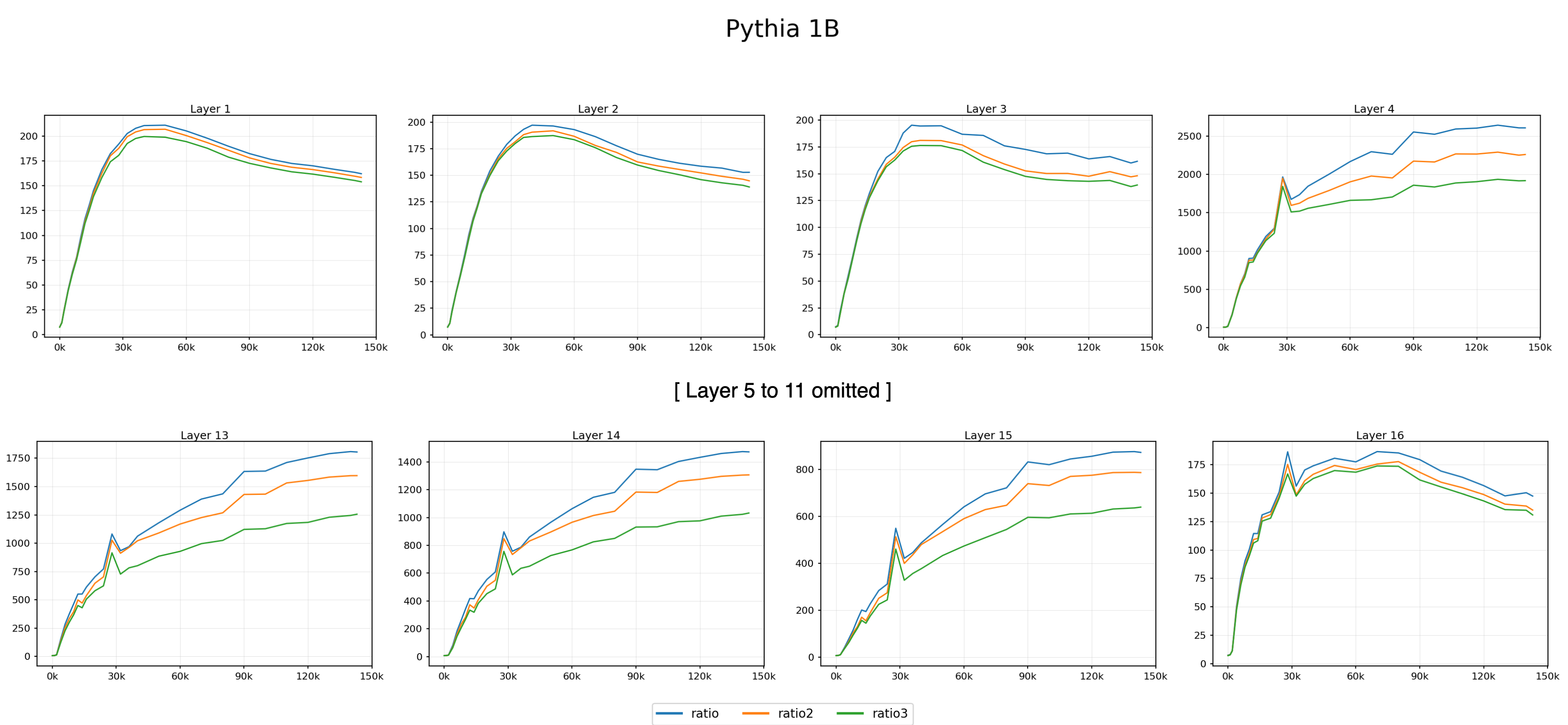}
    \caption{Evolution of the ratio of top activations to median (Equation \ref{eq:ratio}) during training for Pythia 1B. It is a linear interpolation of 37 data points corresponding to different training checkpoints. Apart from the highest activation which is the focus of our study, we also plot ratios corresponding to the top 2 and 3 for comparison. The plots show the training steps on the x-axis, and the ratio of the top magnitudes to median activations on the y-axis.}
    \label{fig:1b-evolution-ratios-per-layer}
\end{figure}
\par
Massive activations are learned throughout training, as they are not present at model initialization (see Figure \ref{fig:ma_at_start_end}), which motivates our work in discovering exactly when and how they develop. We thus plot how they evolve during training and analyze the resulting data, see Figure \ref{fig:1b-evolution-ratios-per-layer}.

\begin{samepage}
\noindent We discover several clear patterns:\begin{enumerate}
    \item \textbf{Layer differentiation} - The evolution of shallow, middle, and deep layers exhibit starkly different shapes.
    \item \textbf{Strong Predictability} - Experiments show we can predict MA evolution accurately by fitting an exponentially decaying, log-modulated function, scoring an average coefficient of determination of 0.984.
    \item \textbf{Stage-wise Development} - MAs often peak early on during training and then monotonically decrease from there, showcasing two clear stages.
\end{enumerate}

\end{samepage}

\subsubsection*{Layer differentiation}

\begin{figure}[h]
  \centering
  \begin{subfigure}[c]{0.45\textwidth}
    \centering
    \includegraphics[width=\linewidth]{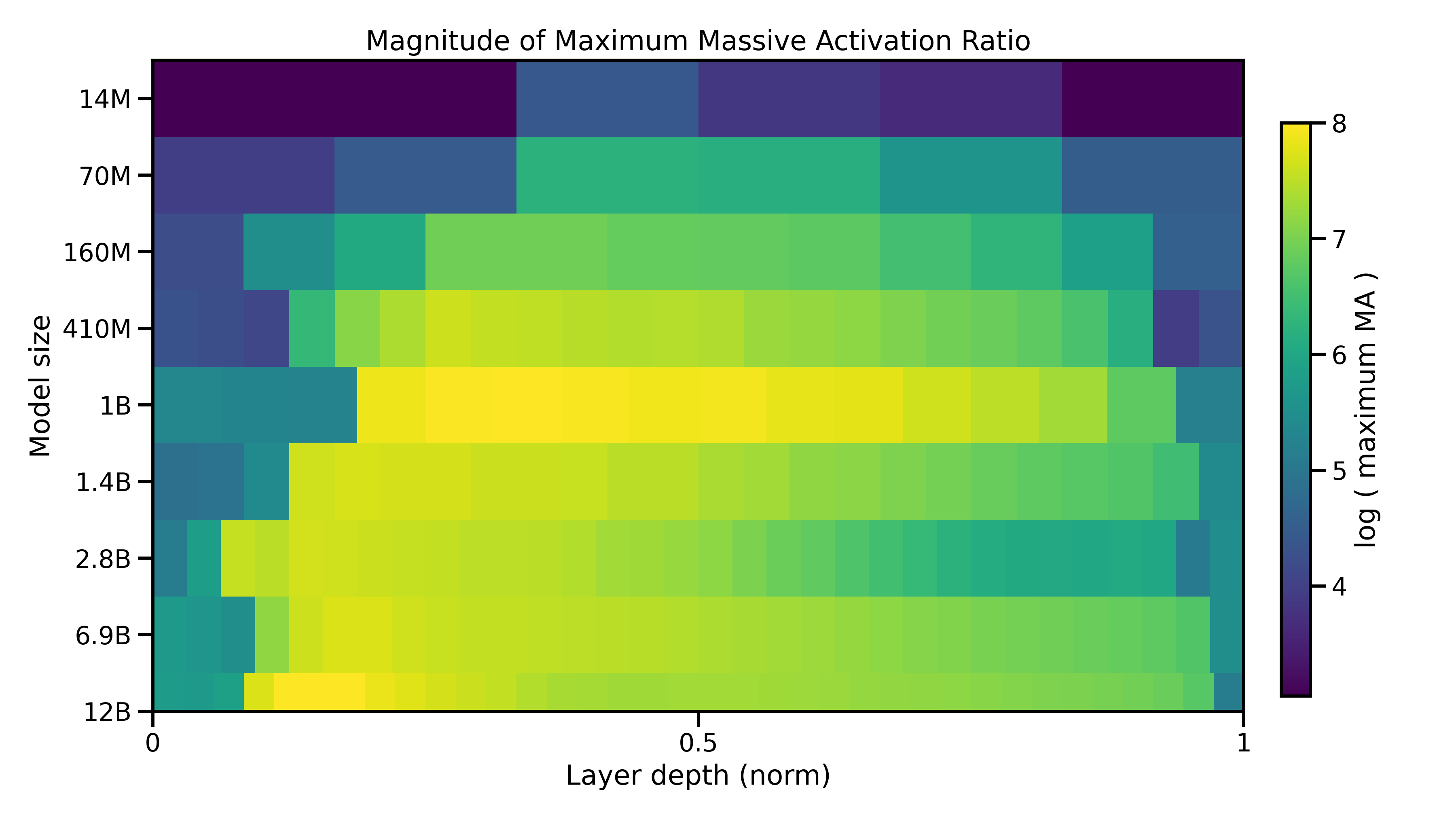}
    \caption{Middle-depth layers display significantly higher MAs than shallow and deep layers.}
    \label{fig:peak-analysis-y}
  \end{subfigure}
  \quad
  \begin{subfigure}[c]{0.45\textwidth}
    \centering
    \includegraphics[width=\linewidth]{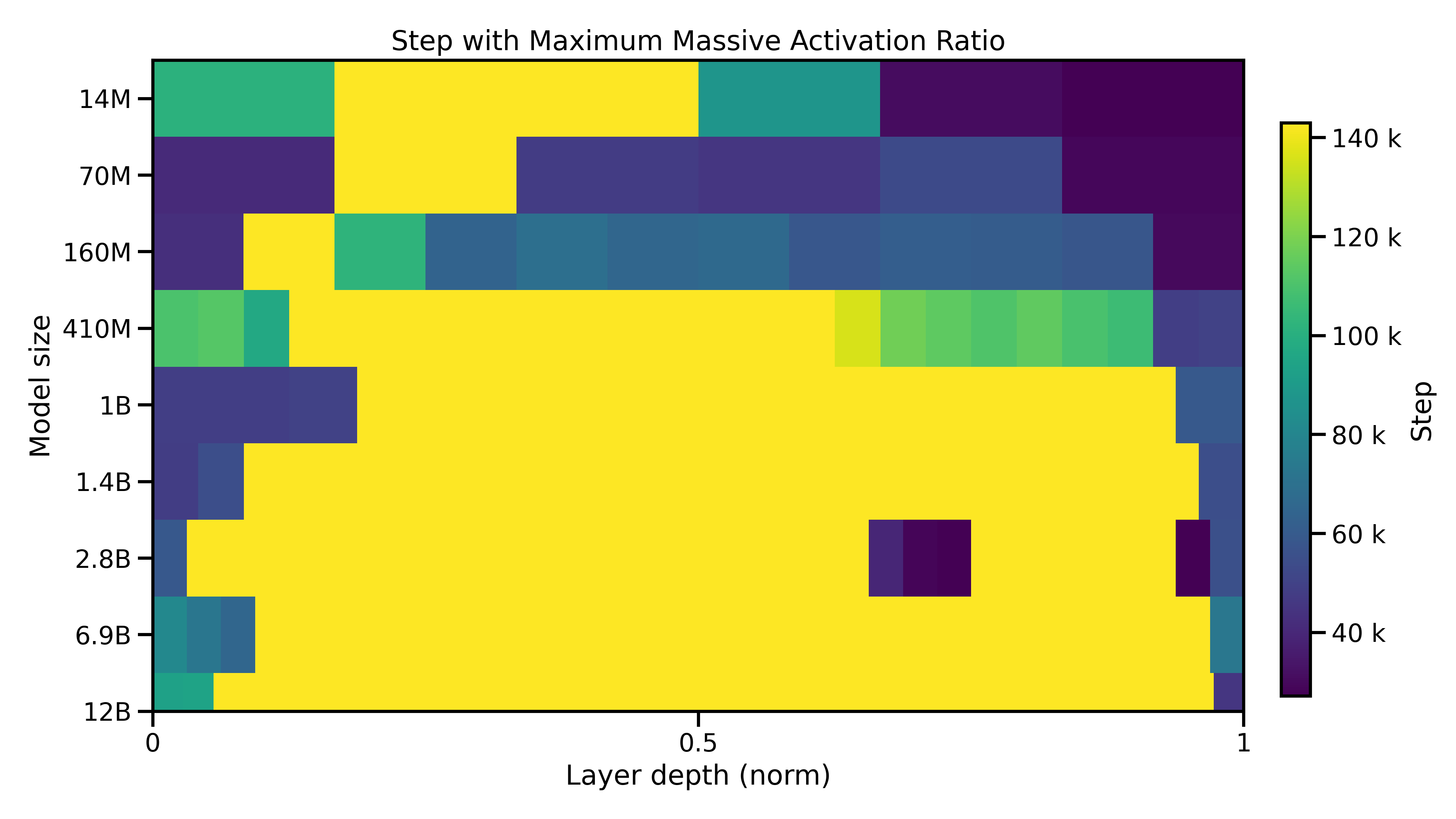}
    \caption{Note the very stark change from shallow and deep layers to middle ones, particularly in the bigger (>410M) models.}
    \label{fig:peak-analysis-x}
  \end{subfigure}
    \label{fig:peak-analysis}
    \caption{Heatmaps showing the location and magnitude of peak MAs by layer depth and model size. Training ends at 143k for the Pythia family, so the yellow middle layers in \ref{fig:peak-analysis-x}  show that MAs would continue to rise monotonically if training continued, where as the darker layers, generally shallow and deep layers, peak and start decreasing before training ends.}
\end{figure}

In terms of MA \textit{magnitude}, the first few shallow and last few deep layers overall have significantly smaller MAs than the middle layers. Observing Figure \ref{fig:peak-analysis-y}, we note that systematically, between 1 and 3 shallow layers, and 1 or 2 deep layers have significantly different MA patterns than the rest of the layers. We observe that bigger models tend to support this pattern more strongly, with the exception of size 2.8B that has noisier data. This is seen in Figure \ref{fig:r2-heatmap} where we observe bigger models have less noisy MA trajectories that can be modeled with Equation \ref{eq:lognorm} with higher confidence.

In terms of MA \textit{temporal dynamics}, most model sizes and layers display smooth MA trajectories with very similar shapes, with only a reduced number of model sizes and layers displaying noisier time series. Across all model sizes, we observe two broad classes of MA trajectories, largely determined by layer depth:

\begin{itemize}
    \item \textbf{Early peak}: Shallow and deep layers exhibit a rapid rise, reach a clear maximum early in training, then decay toward an asymptote.  
    \item \textbf{Log increase}: Middle layers follow a smooth logarithmic climb with no apparent peak during the training window.
\end{itemize}

These patterns are exemplified in Figure \ref{fig:1b-evolution-ratios-per-layer}, which shows layers 1, 2, 3 and 16 to follow an “early peak” pattern, with layers 4 to 15 displaying a “log increase” curve. More systematically, Figure \ref{fig:peak-analysis-x} shows the stark change in pattern in early and late, that peak during training vs middle layers, that peak at step 143k, which is the end of training. Middle layers peak at the end of training as they are monotonically increasing, do to their logarithmic shape. This pattern is particularly stronger for larger models, with 410M or more parameters.

\subsubsection*{Strong Predictability}

Based on our observations of the MA trajectories, we sought to find a low-dimensional functional form hypothesis that could describe the two main observed dynamics: a logarithmic shape, and an initial peak with eventual decay. We evaluated a suite of low-parameter functions to eventually find the function we present in this section: an exponentially decaying, log-modulated function. The theoretical justification of this exact mathematical pattern remains an open question that our empirical findings now motivate.

\begin{equation}\label{eq:lognorm}
f(t)
  = A\,e^{-\lambda x_t}\,
    \log\!\text{(}x_t\text{)}
    + K,
  \qquad
  \text{ where } \;\;\;
  x_t = \gamma t + t_0
\end{equation}

\noindent This unified model fits both the “early peak” and “log increase” regimes across all model sizes and depths with high fidelity, by adjusting the influence of decay with the $\lambda$ parameter, which can make the curve purely logarithmic when $\lambda =0$, or decaying if $\lambda \gg 0$.

Equation \ref{eq:lognorm} forms the core of our following analysis. The fitting process is described further in the Methodology portion. It has 5 parameters: $\{A, \lambda, \gamma, t_0, K\}$, where $A$ controls the amplitude, $\lambda$ the decay rate, $\gamma$ the time scaling, $t_0$ the time offset, and $K$ the asymptotic baseline. The proposed equation was based upon the observation of the massive activation trajectories of each layer, and seeks to unify all models and layers under a single general formula. Note that for each model and each layer, these 5 parameters take different values. The dynamics of the MA trajectories are very similar across models and layers, but their magnitudes and curves vary from model to model.

\begin{table}[h]
\centering
\begin{tabular}{lccccccccc}
\toprule
Model size & 14M & 70M & 160M & 410M & 1B & 1.4B & 2.8B & 6.9B & 12B \\
\midrule
$R^2$      & 0.9307 & 0.9681 & 0.9831 & 0.9937 & 0.9922 & 0.9956 & 0.9686 & 0.9954 & 0.9829 \\
\bottomrule
\end{tabular}
\caption{Average layer-wise best $R^2$ scores — quality of fit of Equation \ref{eq:lognorm} to the time series in Equation \ref{eq:ratio} — for each Pythia model size.}
\label{tab:avg_best_r2}
\end{table}

\begin{figure}[h]
    \centering
    \includegraphics[width=0.95\linewidth]{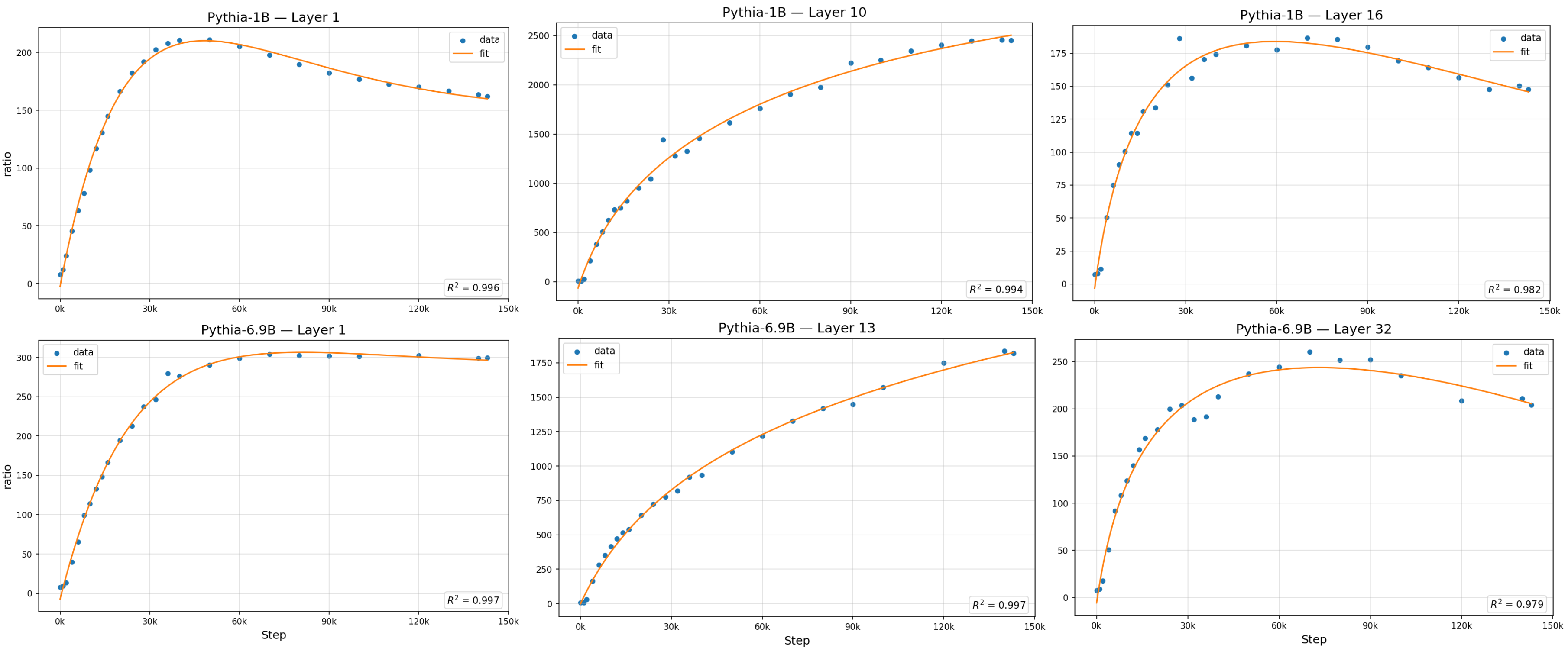}
    \caption{Example fits for two model sizes: 1B and 6.9B and an example shallow, middle and deep layer from each. The plot shows the best fit for Equation \ref{eq:lognorm}, and data points corresponding to Equation \ref{eq:ratio}. The last training step for the Pythia family is 143k.}
    \label{fig:example-fits}
\end{figure}

\begin{figure}[h]

  \centering
  \begin{subfigure}[c]{0.3\textwidth}
    \centering
    \includegraphics[width=\linewidth]{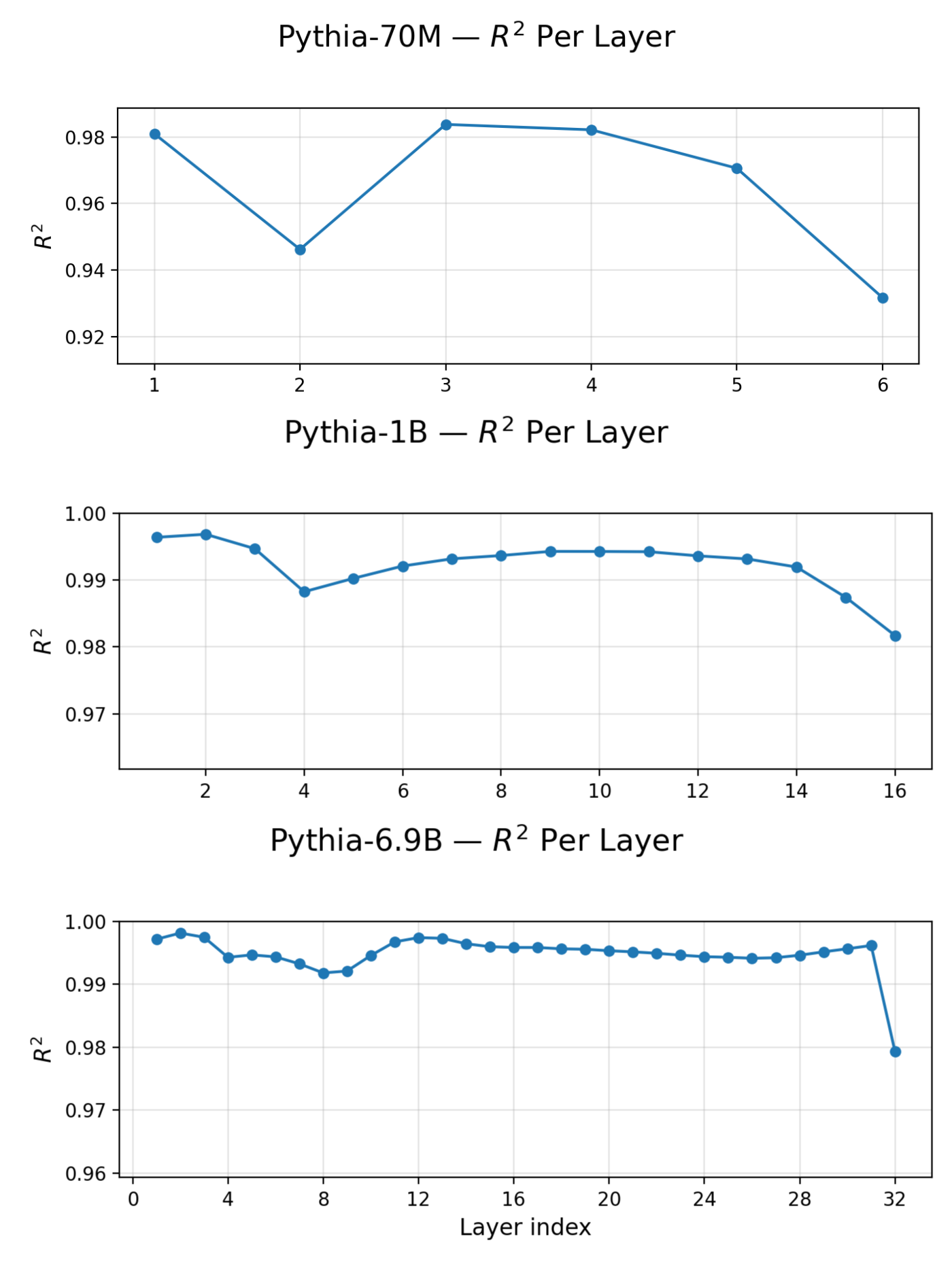}
    \caption{Example $R^2$ values for various models.}
    \label{fig:r2-vertical-plot}
  \end{subfigure}
  \quad
  \begin{subfigure}[c]{0.6\textwidth}
    \centering
    \includegraphics[width=\linewidth]{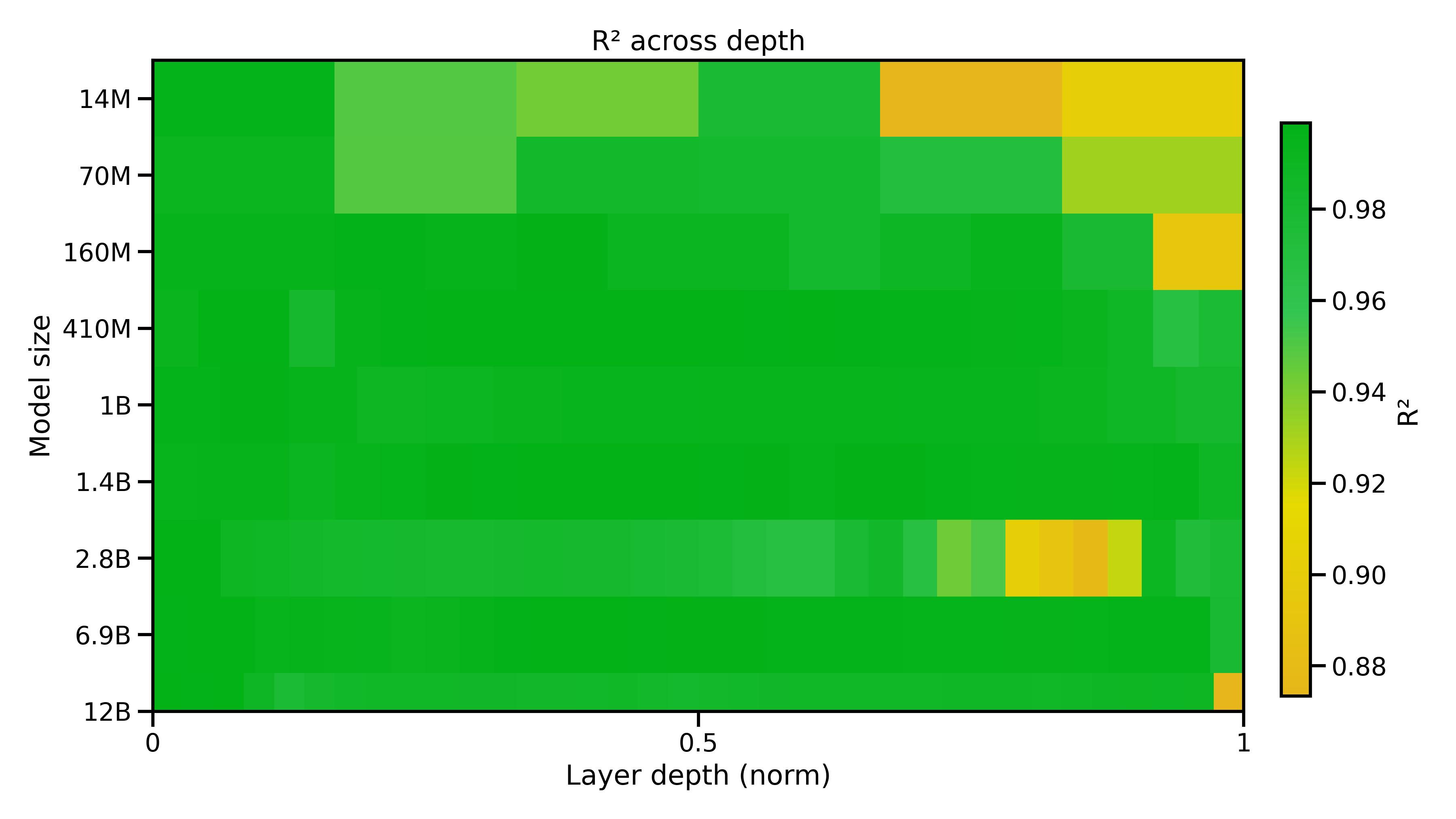}
    \caption{Full map of all models and all layers and the strength or ease of fit. A greener color means that massive activation trajectories in that coordinate can be modeled with high confidence with Equation \ref{eq:lognorm}. Even the lower scoring locations still show evidence of reasonable fits.}
    \label{fig:r2-heatmap}
  \end{subfigure}
  \caption{Coefficients of Determination for the MA trajectory fits.}
\end{figure}

Our proposed log-modulated exponential model predicts the MA evolution with outstanding accuracy, achieving a mean coefficient of determination of 0.984 across 188 layers from nine model sizes\footnote{Average taken over all fitted layers.}. Table \ref{tab:avg_best_r2} reports the average $R^2$ for each model. Notably, smaller models (14M \& 70M) already reach $R^2>0.93$, while larger sizes (160M and above) are generally above $R^2=0.98$, indicating that the MA pattern becomes even more pronounced and regular as model size grows.

Figure \ref{fig:example-fits} showcases representative fits for shallow, middle, and deep layers in both Pythia 1B and Pythia 6.9B, illustrating how the same exponentially decaying log function captures the full range of training dynamics. In turn, Figure \ref{fig:r2-heatmap} breaks down the full layer-wise $R^2$ distribution across all nine sizes, revealing that—even at the extremes—no layer falls below $R^2\approx0.93$, and most cluster tightly around $R^2\approx0.99$. This robustness underlines the universality of our fitted form across depth and scale.

The strength of these results is two-fold. Firstly, the magnitude we are modeling is a result of LLM training - a very noisy process - so an average $R^2$ value greater than 0.98 is surprising. Secondly, the curve hypothesis is a very simple 5-parameter model, which is able to generalize to 9 model sizes and 188 layers.

The activations we are modeling vary with respect to different input sequences, which is why we construct our data points from an average over 10 input sequences. We report on the noise calculated over the 10 data points, and conclude the sample size is sufficient, giving us an average standard error of the mean percentage of 2 points.

\begin{table}[]
    \centering
    \caption{Report of various statistical metrics to quantify variability in the activation ratios we measure, calculated over the 10 samples that were used as input sequences. The metrics are Coefficient of Variation percentage, Standard Error of the Mean percentage, and 95\% Confidence Intervals.}
    \begin{tabular}{l l r r r r}
        \toprule
        Model & CV\% & SEM\% & 95\% CI $\pm$\% \\
        \midrule
              Pythia-14M & 6.95 & 2.20 & 4.97 \\
              Pythia-70M & 7.00 & 2.21 & 5.00 \\
             Pythia-160M & 6.98 & 2.21 & 4.99 \\
             Pythia-410M & 4.64 & 1.47 & 3.32 \\
               Pythia-1B & 6.57 & 2.08 & 4.70 \\
             Pythia-1.4B & 5.19 & 1.64 & 3.71 \\
             Pythia-2.8B & 6.46 & 2.04 & 4.62 \\
             Pythia-6.9B & 5.78 & 1.83 & 4.13 \\
              Pythia-12B & 4.22 & 1.34 & 3.02 \\
        \bottomrule
    \end{tabular}
\end{table}

\subsubsection*{Stage-wise Development}

We have found evidence that in early and late layers across all model sizes, MAs quickly develop in the first ~60k steps before starting to monotonically decrease. This occurs in every model and layer that is colored anything but yellow in Figure \ref{fig:peak-analysis-x}. To the best of our knowledge, this is the first time this phenomenon has been recorded, and it opens up an exciting path of deeper understanding of LLMs. The critical point at which MAs start decreasing suggests there is an underlying two-stage development process governing LLM learning that is poorly understood. Proof of the existence and uniqueness of this critical point is discussed in the next section.

We hope this discovery motivates future work to understand these developmental stages in the field of mechanistic interpretability focusing on \emph{training-dynamics}.

\subsection*{Predicting Massive Activation Trajectories from Transformer Architecture} 

Parameters within Equation \ref{eq:lognorm} are highly influential in the overall shape and size of the curve of massive activation ratios (Equation \ref{eq:ratio}). We now analyze the mathematical behavior of the equation itself. Let us look at steady-state behavior of the equation. In the limit of $t \longrightarrow \infty$, the equation reduces via L'Hôpital's rule for indeterminate forms:
\begin{equation}
    \lim_{t \to \infty} f(t) = \lim_{t \to \infty} Ae^{-\lambda(\gamma t + t_0)}\log(\gamma t + t_0) + K = K
\end{equation}
So the parameter $K$ can be seen to be related to the final steady state value of the ratio. Indeed, in the limit as train steps go to infinity it mathematically is the steady state value, however in practice it is also impacted by the value of the exponential term. Clearly the $A$ parameter affects the overall height of the peak, but let's investigate where the peak occurs. We can do this by first taking the derivative of Equation \ref{eq:lognorm}:
\begin{equation}
    \frac{d}{dt}f(t) = A\gamma e^{-\lambda(\gamma t + t_0)}\left[\frac{1}{\gamma t + t_0} - \lambda\log(\gamma t + t_0)\right]
\end{equation}
Setting this derivative to zero gives us:
\begin{equation}\label{eq:lambertp0}
    \frac{1}{\gamma t + t_0} = \lambda\log(\gamma t + t_0)
\end{equation}

\begin{figure}[h]
    \centering
    \includegraphics[width=1\linewidth]{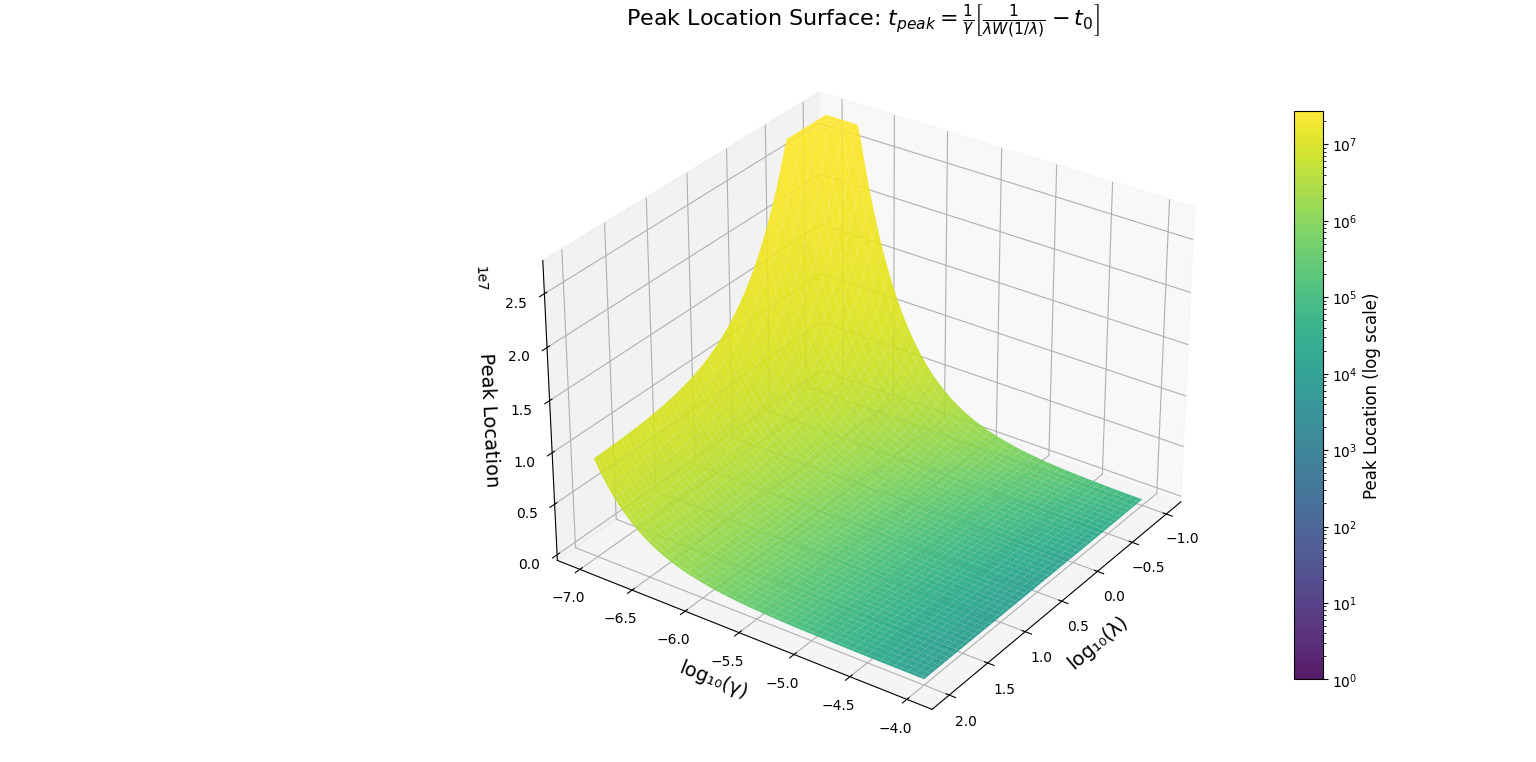}
    \caption{Surface plot illustrating peak location $t_{peak}$ as a function of parameters $\lambda$ and $\gamma$ from Equation \ref{eq:lambert}, with $t_0$ fixed at a typical value found across the Pythia model family. Observable peaks require $t_{peak} > 0$, i.e., $\frac{1}{\lambda W(1/\lambda)} > t_0$; regions violating this constraint are masked. The inverse dependence on $\gamma$ means small changes in this parameter produce large shifts in peak timing.}
    \label{fig:lambert_function}
\end{figure}

Rearranging this critical point equation yields $1 = \lambda (\gamma t + t_0) \log(\gamma t + t_0)$, which can be solved exactly using the Lambert W function. The solution is:
\begin{equation}\label{eq:lambert}
    t_{peak} = \frac{e^{W(1/\lambda)} - t_0}{\gamma}
\end{equation}
where $W$ is the Lambert W function. This analytical solution reveals several key insights about peak behavior. For any positive decay rate $\lambda > 0$, a real-valued critical point always exists since $W(1/\lambda)$ is defined for all positive arguments. However, an observable peak during training requires $t_{peak} > 0$, which imposes the constraint $\frac{1}{\lambda W(1/\lambda)} > t_0$. When this condition is not satisfied, the peak occurs before training begins ($t < 0$), and only monotonic decay is observed. Second, the peak location scales inversely with $\gamma$ (smaller $\gamma$ shifts the peak to much larger training steps) and is offset by $t_0$. This mathematical analysis suggests that $\gamma$ and $\lambda$ are the most critical parameters for controlling curve shape and location, with $\gamma$ having particularly strong influence on peak timing (shown in Figure \ref{fig:lambert_function}) due to its position in the denominator. Figure \ref{fig:lambert_function} illustrates the relationship between the number of training steps needed to see a peak in Equation \ref{eq:ratio} and parameters $\gamma$ and $\lambda$. Note that the Pythia model family did not reach training steps greater than $\approx 143k$, so for any fitted equation parameters that predict a peak at location greater than the maximum number of training steps, a peak would not have been seen for a particular layer during the training cycle. As seen in our previous results section, certain layers within each model exhibit monotonically increasing ratio (\ref{eq:ratio}) behavior as training progresses, and some layers exhibit a sharp peak early in training and then decay to a steady state value (see Figure \ref{fig:example-fits}). Overall, the layer depth (along with training length) is correlated with peak observation, but we notice that there are breaks in this pattern. For example, the layers (or more specifically, the depth of a layer within a model relative to its overall size) in which quantity \ref{eq:ratio} exhibits a peak is different for Pythia 12B vs Pythia 1B.

The predictability analysis presented below uncovers a relationship between transformer architecture and massive activation dynamics. We demonstrate that the complex, seemingly chaotic emergence of massive activations during training is actually governed by predictable mathematical relationships that can be controlled through architectural design. Specifically, we show how architectural parameters play a role in whether peaks in Equation \ref{eq:ratio} will emerge, their magnitude, and their steady-state behavior—enabling practitioners to architect models with desired massive activation properties from the outset for both further study of massive activations and potentially optimized training dynamics. For instance, controlling $\gamma$ could theoretically allow peaks to occur sooner in the training cycle, allowing steady states to be reached sooner. However, the relationship between massive activation timing and overall training efficiency requires further investigation. Note that the $\lambda$ parameter does effectively control whether a peak exists, but in practice within the Pythia model family, $\lambda$ always takes a value which allows peak existence, therefore $\gamma$ or $t_0$ would play the largest roles in peak observation.

\begin{table}[ht]
\centering
\caption{Machine Learning Model Performance for Predicting Massive Activation Parameters. Values show test set $R^2$ scores. Best performing model for each parameter is shown in bold. Negative $R^2$ values indicate worse-than-baseline performance. Dataset: 188 samples (80\% train, 20\% test), 5-fold cross-validation.}
\label{tab:model_performance}
\begin{tabular}{lcccccc}
\toprule
\textbf{Parameter} & \textbf{Transform} & \textbf{Ridge} & \textbf{Lasso} & \textbf{Random Forest} & \textbf{Gradient Boosting} & \textbf{XGBoost} \\
\midrule
A (Amplitude) & log1p & 0.077 & 0.195 & \textbf{0.476} & 0.274 & 0.244 \\
$\lambda$ (Peak Occurrence) & log1p & 0.031 & -0.015 & 0.643 & 0.506 & \textbf{0.664} \\
$\gamma$ (Peak Location) & log1p & \textbf{0.056} & -0.006 & 0.055 & -1.571 & -0.089 \\
$t_0$ (Time Offset) & log1p & 0.100 & 0.017 & \textbf{0.447} & 0.266 & 0.387 \\
K (Asymptotic Baseline) & yeo-johnson & 0.405 & 0.316 & 0.803 & 0.824 & \textbf{0.847} \\
\bottomrule
\end{tabular}
\end{table}

The fitted models in the predictability analysis showed strong fit characteristics for certain variables (see Table \ref{tab:model_performance}), such as $\lambda$, and $K$, while the variables $A$, $\gamma$ and $t_0$ showed a slightly weaker performance across all models. Table \ref{tab:model_performance} shows the performance of various standard machine learning algorithms in predicting the values of parameters in Equation \ref{eq:lognorm} using only features constructed using architectural specifications for various models within the Pythia family. In most cases we fit the models to transformed features due to large outliers which can easily skew predictions for most model types. Additionally, target variables were isolated such that other variables in the equation were not used as predictors to help isolate the effects of the architecture from the fitted equation. Below, we offer an in-depth analysis of key features and their fits, as well as key explainability measures the top models (shown in bold in Table \ref{tab:model_performance}) were able to determine using a SHAP and PDP feature analysis. This interpretability analysis gives us a way to control the value of the behaviors of our MA ratio (Equation \ref{eq:ratio}) using purely architectural design patterns. We focus on parameters $\lambda$ and $K$ due to their strong fits. 


$A$ and $t_0$ form amplitudes and fixed offsets of the peak and are less impactful and have lower $R^2$ scores, so we omit their discussion, along with $\gamma$ due to its lower performance, for brevity. Table \ref{tab:feature_definitions} provides interpretable definitions of various features used in the predictive ML models. Note that raw features are not used to avoid having direct dependency on specific models with the Pythia family; we primarily focus on normalized features to increase applicability more widely across the model family, in general. We also note that the architectural design choices within the Pythia model family do not have high degrees of variation, and often a particular feature scales proportionally to model size. So machine learning models fit using this architectural distribution could have trouble generalizing.

\begin{table}[ht]
\centering
\caption{Key Architectural Features Used in Predictive Models. \textbf{Note:} $\ell$ = layer index, $L$ = total layers, $d$ = hidden dimension, $H$ = attention heads, $d_{ff}$ = intermediate (MLP) dimension. }
\label{tab:feature_definitions}
\begin{tabular}{p{4cm}p{3cm}p{6cm}}
\toprule
\textbf{Feature Name} & \textbf{Formula} & \textbf{Interpretation} \\
\midrule
Layer Position & $\ell/L$ & Relative depth within model (0 = first layer, 1 = last layer) \\
Layer Position² & $(\ell/L)^2$ & Quadratic position effects for non-linear layer behavior \\
Layer Position³ & $(\ell/L)^3$ & Cubic position effects capturing complex depth dependencies \\
Layer Position$^{1/2}$ & $\sqrt{\ell/L}$ & Square root position effects for early-layer emphasis \\
\midrule
Attn. Heads/Hidden Size & $H/d$ & Number of attention heads per hidden dimension \\
Intermediate Ratio & $d_{ff}/d$ & MLP expansion factor (in the Pythia family this a fixed value of 4x) \\
Width/Depth Ratio & $d/L$ & Model width relative to depth (architectural shape) \\
Attn. Heads/Num. Layers & $H/L$ & Attention head budget distributed across layers \\
\midrule
log(Hidden Size) & $\log(d)$ & Logarithm of hidden dimension (proxy for model size) \\
Layer DepthxModel Depth & $\ell L$ & Interaction between layer location and total length \\
\bottomrule
\end{tabular}
\end{table}

\subsubsection*{Parameter $K$}

Parameter $K$ represents the steady-state value of massive activation ratios in the limit as training steps approach infinity, making it a critical architectural design target. The analysis reveals that this asymptotic behavior is highly predictable from architectural choices, with the model achieving an $R^2$ of $0.8186$ on the test set.

\begin{figure}[H]
    \centering
    \begin{subfigure}[t]{0.48\textwidth}
        \centering
        \includegraphics[width=\linewidth]{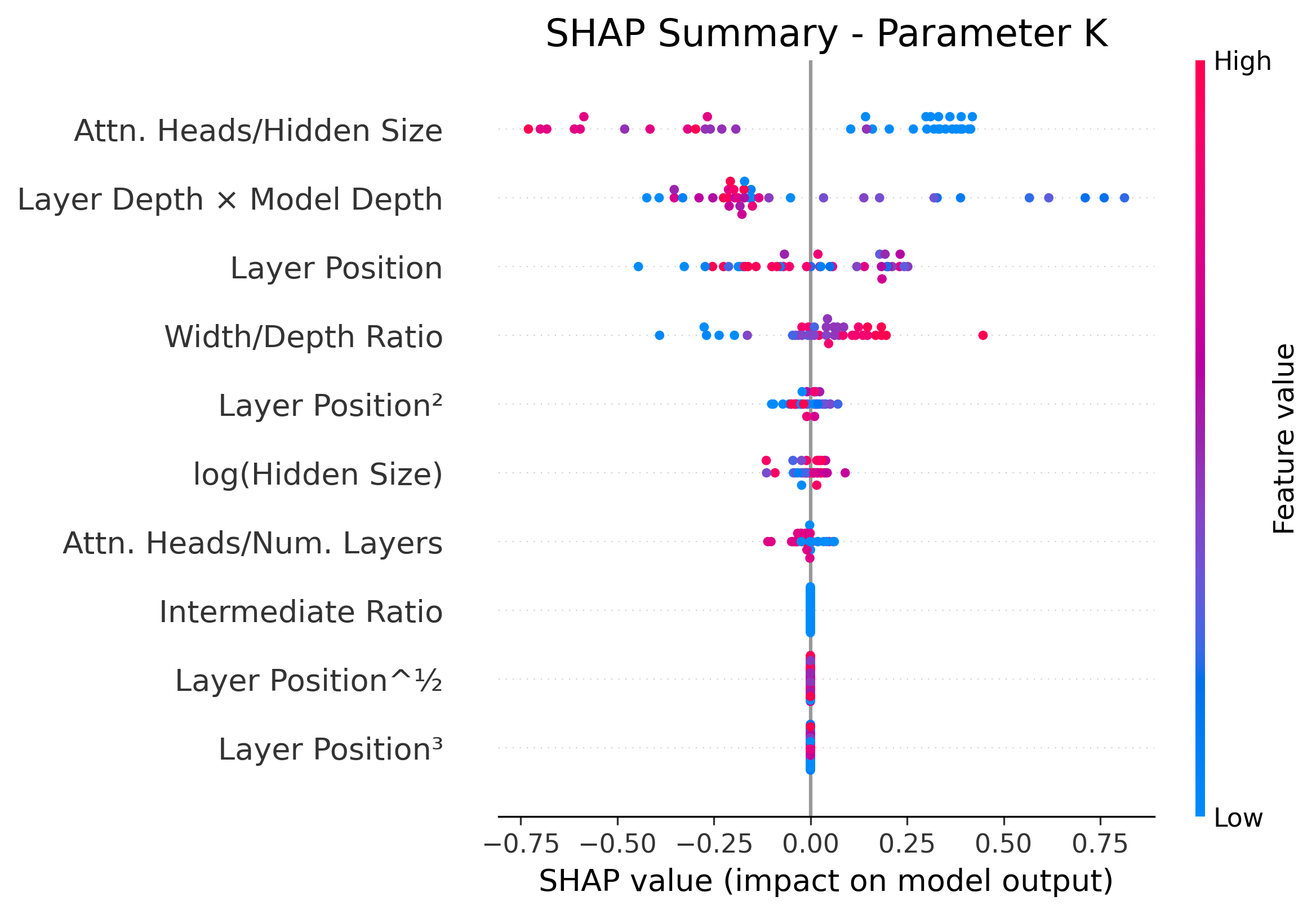}
        \caption{SHAP Summary Plot for Parameter K revealing directional relationships between architectural features and steady-state predictions. Colors indicate feature values (red = high, blue = low).}
        \label{fig:k_shap_summary}
    \end{subfigure}
    \hfill
    \begin{subfigure}[t]{0.48\textwidth}
            \centering
        \includegraphics[width=\linewidth]{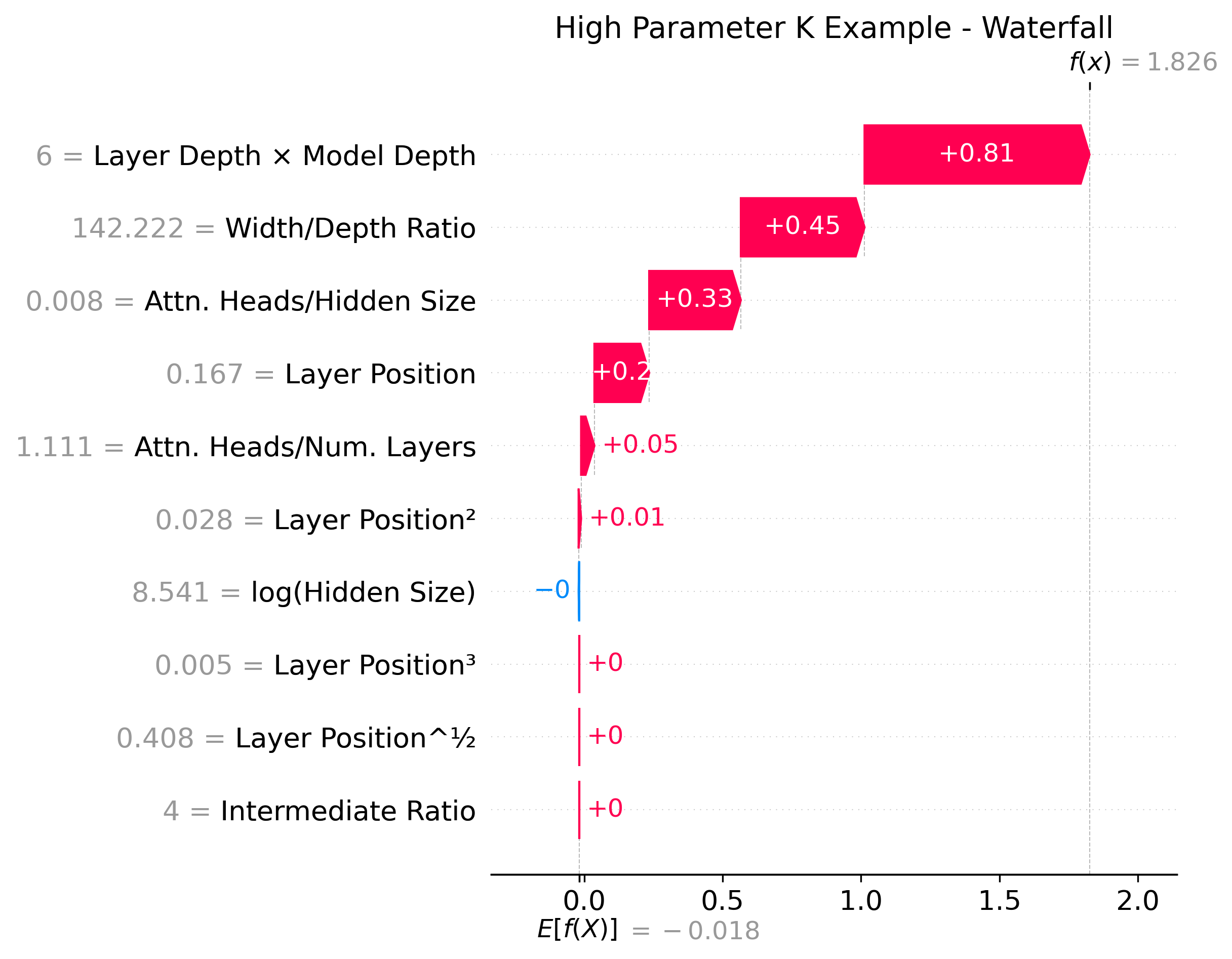}
        \caption{SHAP Waterfall Plot for highest Parameter K prediction showing individual feature contributions. Predicted value f(x) = 1.811 in transformed space corresponds to approximately 5.1 in original scale.}
        \label{fig:k_shap_waterfall}
    \end{subfigure}
    \caption{}
\end{figure}

The SHAP feature importance analysis identifies attention density (attention heads per hidden dimension) as the dominant architectural control for steady-state behavior, followed by the layer depth interaction term and layer position (as seen in Figure~\ref{fig:k_shap_summary}). This hierarchy indicates that attention architecture design has the strongest influence on long-term activation dynamics, while layer-specific effects provide secondary modulation.

The SHAP summary plot in Figure~\ref{fig:k_shap_summary} reveals the directional relationships governing steady-state control. Most notably, decreasing attention density—either by reducing the number of attention heads or increasing the hidden dimension—consistently increases Parameter
K, leading to higher steady-state massive activation ratios. This relationship suggests that models with fewer, larger attention heads will exhibit elevated baseline activation levels in the long term compared to models with many smaller heads. This finding has practical implications for model architecture design: decisions about head count and hidden dimension size do not merely affect computational efficiency or representational capacity, but also systematically influence the long-term magnitude of massive activations within the network.

The waterfall plot in Figure~\ref{fig:k_shap_waterfall} demonstrates these effects in practice, showing how a high Parameter $K$ prediction results from specific architectural choices. The largest contribution (+0.81) comes from the layer depth interaction term, indicating this is a deep layer within a deep model. Additional positive contributions from low attention density (+0.33) and high width/depth ratio (+0.45) further elevate the steady-state prediction, while early layer position (+0.01) provides additional upward pressure.

These findings establish that transformer architects can systematically control steady-state massive activation behavior through attention architecture design, with attention density serving as the primary control mechanism and layer depth providing amplification effects for deeper models.

\subsubsection*{Parameter $\lambda$}

Lastly, we provide analysis for parameter $\lambda$, which influences whether peaks are observable during training. For any $\lambda > 0$, a mathematical peak exists, but observability requires $t_{peak} > 0$, i.e., $\frac{1}{\lambda W(1/\lambda)} > t_0$. Since larger $\lambda$ values decrease the critical point $x_t^* = \frac{1}{\lambda W(1/\lambda)}$, high $\lambda$ makes it more likely that the peak occurs before training begins ($t_{peak} < 0$), rendering it unobservable.

\begin{figure}[H]
    \centering
    \includegraphics[width=.8\linewidth]{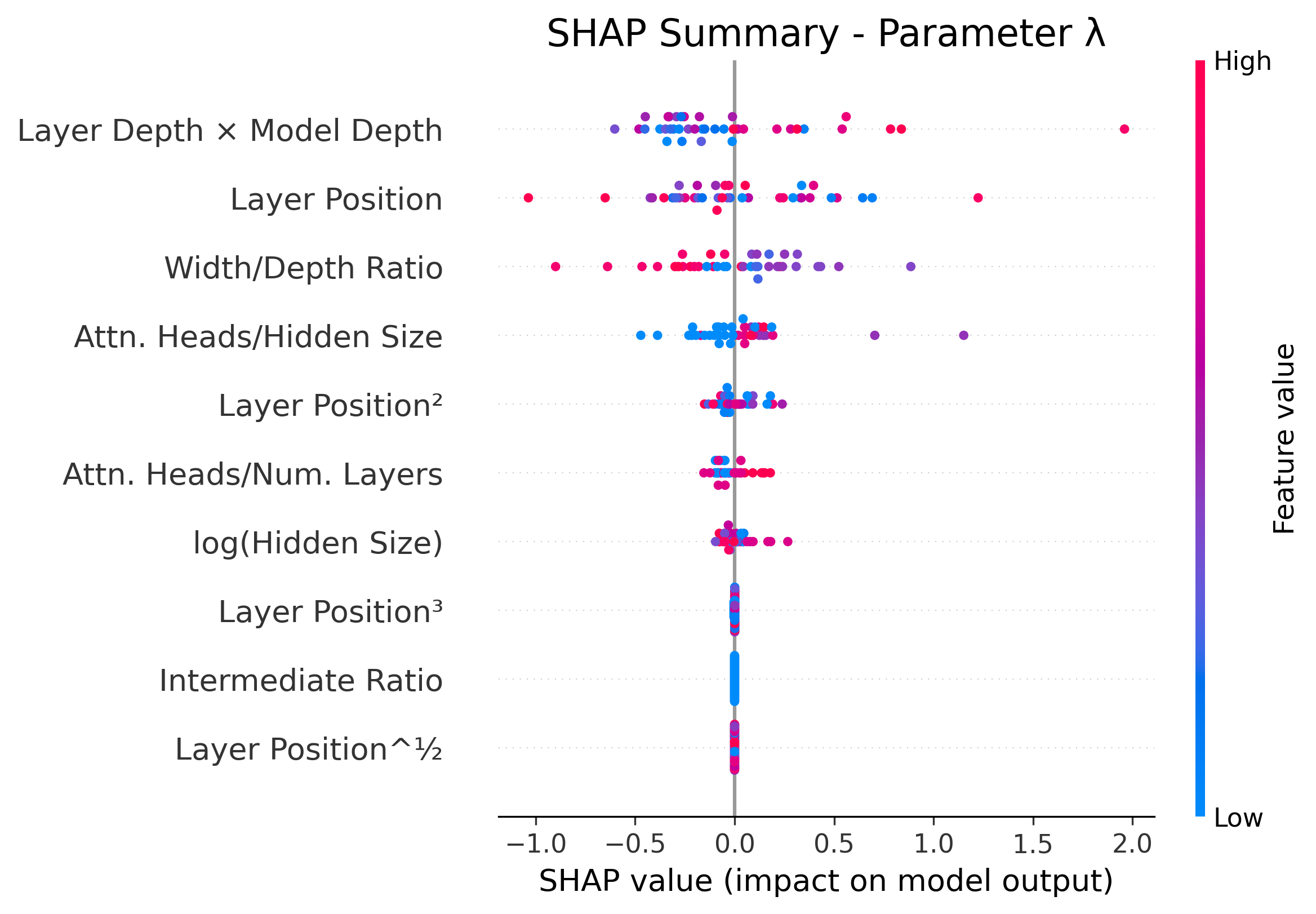}
    \caption{SHAP summary plot for parameter $\lambda$, indicating directional relationships between architectural choices and values of $\lambda$ predicted by the top performing machine learning model.}
    \label{fig:lambda_shap_summary}
\end{figure}

The SHAP analysis in Figure~\ref{fig:lambda_shap_summary} reveals that layer depth characteristics dominate $\lambda$ predictions, with the layer depth interaction term showing the strongest and most consistent effects. Deeper layers within deeper models (high feature values, red points) consistently push $\lambda$ toward higher values, effectively suppressing peak behavior. This relationship suggests that peak occurrence follows a systematic architectural pattern, with early layers in shallow models most likely to exhibit peaks, while deeper layers in larger models tend toward monotonic decay. This is exactly what is seen, for example, in \ref{fig:1b-evolution-ratios-per-layer}.


These findings establish a clear architectural hierarchy for controlling peak occurrence: layer depth and model shape provide the primary controls, while attention architecture offers secondary modulation. The high predictive accuracy suggests that architects can systematically design models to either encourage or suppress massive activation peaks across different layers, providing a new dimension of control over internal model dynamics during training.

\section*{Discussion}

This study presents a new framework for understanding the emergence and dynamics of MAs in transformer models. Our core contributions can be summarized as follows. First, we introduce a five-parameter, exponentially-decaying log-modulated function that accurately models how the top-magnitude activations evolve throughout training within each transformer layer. Each parameter: amplitude ($A$), decay rate ($\lambda$), time scaling ($\gamma$), time offset ($t_0$), and asymptotic baseline ($K$), captures an interpretable aspect of the activation emergence curve. This mathematical model provides, for the first time, a quantitative and interpretable description of the temporal development of MAs across model scales and layers.
\par
Second, we develop a machine learning-based predictive framework that can forecast the values of these five parameters solely from architectural features such as the number of layers, hidden dimension size, attention head count, and layer position. By leveraging tree-based ensemble models (Random Forest~\cite{breiman2001random}, XGBoost~\cite{chen2016xgboost}) and modern explainability techniques (SHAP values~\cite{lundberg2017unified}, partial dependence plots~\cite{friedman2001greedy}), we show that it is possible to predict, with high accuracy, key aspects of massive activation dynamics---notably the steady-state baseline $K$ ($R^2=0.847$) and curvature features $\lambda$ ---based purely on a model's static architecture. This demonstrates that MAs are not random artifacts of training, but follow systematic, architecture-dependent rules.
\par
Third, our interpretability analyses reveal that specific architectural choices serve as master controls for MA dynamics. In particular, the ratio of attention heads to hidden dimension (``Attn. Heads/Hidden Size") and layer position (``Attn. Heads/Num. Layers") within the network are the dominant drivers of MA emergence and amplitude. We identify that changing these ratios within a transformer architecture, for example, changing ``Attn. Heads/Layer" from 0.7 to 1.0, can have direct impact in controlling the shape of the development of MA ratio curves during the training cycle, and change the distribution of which layers have a visible peak and which do not. This non-monotonic relationship suggests that careful tuning of attention architecture offers a concrete handle for modulating high-norm activation behavior. 
\par
Taken together, these findings establish that the emergence of massive activations is a predictable, quantifiable, and interpretable phenomenon rooted in architectural design---not merely a quirk of optimization or dataset. MAs appear abruptly during training, often at specific layers, hidden dimensions, and token positions (e.g., beginning-of-sequence or delimiter tokens \cite{Sun:massiveactivations:2024,owen2025refinedanalysismassiveactivations}), and then stabilize to input-agnostic, nearly constant values that act as implicit bias terms within the network. 
 \par
This new understanding of MA dynamics opens several avenues for both theory and practice. For model designers, our predictive framework enables MA-aware architecture by providing precise control over when and where massive activations emerge. Rather than treating MAs as unpredictable training artifacts, architects can now systematically design models to either suppress or accentuate MAs as desired. For example, our findings show that adjusting attention density can create sharp phase transitions in steady-state behavior, while modifying width/depth ratios provides coordinated control over both peak timing. This offers a principled way to navigate architectural tradeoffs while maintaining desired MA properties.
\par
While our study is comprehensive within the EleutherAI Pythia family of decoder-only transformers, several limitations remain. Our results may not fully generalize to encoder-based models (see BERT, \cite{devlin2018bert}), sequence-to-sequence architectures (see orginal transformer variant outlined in \cite{vaswani2017attention}), or other architectures such as LLaMA-based models \cite{touvron2023llama2openfoundation}. The temporal resolution of our checkpoint data means we capture the emergence of MAs at coarse granularity; more frequent checkpoints might uncover finer dynamics or more gradual onset. Our input sampling for activation measurement was necessarily limited for computational reasons; broader input distributions may reveal additional or rare MAs not captured here. Finally, while our framework robustly predicts several parameters (especially $K$, $\lambda$), timing-related parameters ($\gamma$ and $t_0$), remain harder to predict, suggesting that some aspects of MA emergence depend on optimization dynamics or data ordering not captured in architecture alone.
\par
These limitations point toward several promising directions for future research. First, our observation that some layers require extended training beyond 143k steps to reach their predicted peaks raises intriguing questions about the relationship between MA dynamics and grokking phenomena \cite{power2022grokking,liu2022towards}. Since grokking often occurs after hundreds of thousands of training steps, future work could investigate whether this extended training provides sufficient time for ``slow-peaking" layers to complete their internal reorganization, potentially revealing MA peak timing as a predictor or correlate of delayed learning transitions. Additionally, our predictive framework suggests the possibility of designing quantization-aware architectures that intentionally delay MA peak emergence well beyond typical training horizons. Since many applications require quantization for deployment, architectures that can maintain performance while keeping MAs suppressed during standard training durations could offer significant practical advantages for efficient inference.
\par
Second, validation across diverse model families would strengthen the generalizability of our framework. While Pythia provides an excellent controlled environment, expanding to encoder-decoder architectures ~\cite{vaswani2017attention}, different training objectives \cite{radford2018improving}, and alternative attention mechanisms \cite{liu2021swin,choromanski2020rethinking} would test the universality of our architectural control principles. Of particular interest would be investigating whether our predictive relationships hold across models with different positional encoding schemes \cite{su2024roformer,press2021train}, normalization strategies \cite{ba2016layer, zhang2019rmsnorm} and non-normalization strategies \cite{zhu2025transformers}, or activation functions.
\par
Third, our current analysis is constrained by the limited architectural diversity within existing model families. For example, Pythia models consistently use a 4× MLP expansion ratio, making it impossible to predict how variations in this parameter affect MA dynamics. Future work could involve training custom model families with systematic variations in currently fixed ratios—such as MLP expansion factors ranging from 2× to 8×, or models with heterogeneous attention head configurations across layers. Such experiments would provide the architectural diversity needed to validate and extend our predictive framework to the full space of transformer design choices.
\par
In summary, our results provide the first quantitative, predictive, and interpretable model of massive activation emergence in transformers, with immediate implications for theory, design, and deployment. We hope these findings will serve as a foundation for future MA-aware model development and inspire new techniques for harnessing or controlling this fundamental phenomenon.

\section*{Methods}

In this section, we outline the experimental setup used to both fit a mathematical model to the evolution of massive activations during a training cycle, and also to detail the framework used to extract explainable predictive insights from the model.

\subsection*{Experimental setup}

MAs were estimated by setting $X$ in Equation \ref{eq:approximating-mas} to be a sample of 10 random sequences from the Red Pajama dataset \cite{togethercomputer_redpajama_1Tsample_2023}. This dataset represents data from the training distribution, or more generally, from the target distribution. For each model, we elicited MAs for at least 37 regularly spaced steps.

Parameters for Equation \ref{eq:lognorm} were estimated with SciPy 1.15.2 (\verb|scipy.optimize.curve_fit|; \cite{2020SciPy-NMeth}) using the Trust-Region Reflective algorithm \cite{ColemanLi1996TRF}, with analytic Jacobians supplied and bounds enforced on the $\lambda$ parameter to keep it positive, as negative values produce an exponential curve, much different from the observed trajectories. The \verb|curve_fit| algorithm seeks to find a set of parameters that minimize the fit error, and does so iteratively. To speed up convergence, we provide an analytic Jacobian of our function, and an initial guess. Data points are normalized first, fitted in the normalized space, and then the parameters are scaled back. Each model and layer had a minimum of 27 data points, corresponding to regularly spaced training checkpoints.

For our initial guess, we set $A$ to the response range ($\max y-\min y$) and $K$ to the midrange ($\min y + A/2$), for an appropriate the vertical scale and offset of initial curve. We initialize $\lambda=1/\mathrm{std}(x)$ and $\gamma=1/\mathrm{range}(x)$ to keep the effective input ($\gamma x+t_0$) and decay rate on an $O(1)$ scale, which improves conditioning and avoids overly flat or overly stiff starting dynamics. We use $t_0=1$ as a safe positive shift so $\log(\gamma x+t_0)$ is well-defined at initialization.

We reparametrized Eq. (\ref{eq:lognorm}) to reduce parameter coupling and improve numerical stability under the Trust-Region Reflective solver as
\begin{equation}\label{eq:lognorm_reparam}
f(t)
= e^{-\beta t}\bigg [A_1\log\big(t+\tau_0\big)+A_2\bigg ]+K,
\end{equation}
with the parameter mapping
\[
\beta=\lambda\gamma,\qquad
\tau_0=\frac{t_0}{\gamma},\qquad
A_1=A,e^{-\lambda t_0},\qquad
A_2=A_1\log(\gamma).
\]
This follows from $x_t=\gamma t+t_0=\gamma(t+\tau_0)$, so that $\log(x_t)=\log(\gamma)+\log(t+\tau_0)$.

Other hypothesis were tested, such as \textit{first-} and \textit{second-order step response functions}:

\paragraph{First-order step response}
\begin{equation}
    y(t) = K \left(1 - e^{-t/\tau}\right).
\end{equation}

\paragraph{Second-order step response}\quad\\

\textbf{Underdamped} $(0 \le \zeta < 1)$:
\begin{equation}
    y(t) = K \left[
    1 - \frac{1}{\sqrt{1-\zeta^2}}
    e^{-\zeta \omega_n t}
    \sin\!\left(\omega_d t + \arccos(\zeta)\right)
    \right],
    \qquad
    \omega_d = \omega_n \sqrt{1-\zeta^2}.
\end{equation}

\textbf{Critically damped} $(\zeta = 1)$:
\begin{equation}
    y(t) = K \left[1 - e^{-\omega_n t}\left(1 + \omega_n t\right)\right].
\end{equation}

\textbf{Overdamped} $(\zeta > 1)$:
\begin{equation}
y(t) = K \left[
1 - \left(
\frac{r_2}{r_2 - r_1} e^{r_1 t}
+
\frac{r_1}{r_1 - r_2} e^{r_2 t}
\right)
\right],
\qquad
r_{1,2} = -\omega_n \left(\zeta \mp \sqrt{\zeta^2 - 1}\right).
\end{equation}

\begin{table}[]
    \centering
    \begin{tabular}{l c c c c c c c c c c}

        \toprule
        Model hypothesis & 14M & 70M & 160M & 410M & 1B & 1.4B & 2.8B & 6.9B & 12B \\
        \midrule
        Original & 0.9951 & \textbf{0.9829} & 0.8068 & \textbf{0.9832} & \textbf{0.9922} & \textbf{0.9666} & 0.9928 & 0.9947 & 0.9659 \\
        Reparametrized & \textbf{0.9956} & 0.9825 & \textbf{0.9307} & \textbf{0.9831} & \textbf{0.9922} & 0.9630 & \textbf{0.9936} & \textbf{0.9954} & \textbf{0.9681} \\
        Step 1 & 0.8008 & 0.6177 & -0.6833 & 0.2592 & 0.6183 & 0.3134 & 0.8202 & 0.8774 & -0.3587 \\
        Step 2 & 0.9754 & 0.9682 & 0.6316 & 0.9071 & 0.9760 & 0.9619 & 0.9816 & 0.9774 & 0.8031 \\
        \bottomrule
    \end{tabular}
    \caption{Average $R^2$ over all layers for each Pythia size, for the different hypothesis we tested to model MA evolution behavior.}
    \label{tab:r2-per-hypothesis}
\end{table}

These were discarded due to lower $R^2$ and Akaike Information Criterion (AIC) score \cite{akaike1974aic}. We opted for AIC score because we sought to compare not only accuracy of the models, but also simplicity, when comparing the 5 parameter hypothesis in Equation \ref{eq:lognorm}, versus the 3 parameter second degree step function. The strength of performance from Equation \ref{eq:lognorm} compensated the additional parameters, as we can see in Table \ref{tab:r2-per-hypothesis}, which compares the average $R^2$ score per hypothesis. Original and Reparametrized represent the same hypothesis, and strongly dominate over the other choices.



To facilitate further analysis and reproducibility, all processed activation statistics and fitted parameter values from this stage are included in our released dataset (see Data Availability).

\subsection*{Parameter analysis and architectural relationships}

After fitting our mathematical models to the temporal evolution data, we investigate how the learned parameters relate to architectural properties of the transformer models. This predictive analysis enables us to understand which architectural design choices most strongly influence massive activation emergence patterns, and more importantly, provides us with a way to directly control the training dynamics purely through initial model architectural choices. 

\subsubsection*{Feature engineering and data preparation}

We construct a comprehensive feature set from the architectural specifications of each Pythia model variant, including:

\begin{itemize}
    \item \textbf{Core architectural features}: hidden dimension size ($d$), number of layers ($L$), number of attention heads, feed-forward network width, rotary embedding base, and rotary percentage
    \item \textbf{Derived features}: layer position normalized by total depth, ratios between architectural components (e.g., attention heads per hidden dimension), and interaction terms
    \item \textbf{Polynomial features}: quadratic and cubic terms for layer position to capture non-linear positional effects
    \item \textbf{Logarithmic transformations}: log-scaled versions of large architectural parameters to handle wide dynamic ranges
\end{itemize}

\noindent Given the diverse scales and distributions of our fitted parameters, we apply appropriate transformations to improve model performance. For parameters with positive values and high skewness, we use log transformations. For parameters with mixed signs or extreme outliers, we employ a Yeo-Johnson \cite{yeo2000new} power transformation to achieve approximate normality. Features are transformed with a standard scaler to allow usage across a range of model types; plots showing relationships between targets and features, therefore, often represent the relationship between the transformed target and transformed feature. 

\subsubsection*{Machine learning model selection}

We evaluate multiple regression algorithms to identify the best predictor for each fitted parameter:

\begin{itemize}
    \item \textbf{Linear models}: Ridge and Lasso regression with L2 and L1 regularization respectively
    \item \textbf{Tree-based ensembles}: Random Forest and Gradient Boosting regressors to capture non-linear relationships
    \item \textbf{Advanced boosting}: XGBoost with careful hyperparameter tuning for optimal performance
\end{itemize}

\noindent Model selection is based on 5-fold cross-validation performance, with the final evaluation conducted on a held-out test set (20\% of data). We use coefficient of determination ($R^2$) as our primary metric, supplemented by mean absolute error (MAE) and root mean squared error (RMSE).

\subsubsection*{Model interpretability and feature importance}

To understand which architectural factors most strongly influence massive activation dynamics, we employ multiple interpretability techniques:

\begin{itemize}
    \item \textbf{Feature importance analysis}: For tree-based models, we extract built-in feature importance scores.
    \item \textbf{SHAP (SHapley Additive exPlanations) analysis}: Provides model-agnostic explanations of individual predictions and global feature importance.
    \item \textbf{Partial dependence plots}: Visualize the marginal effect of individual features on predicted parameter values
    \item \textbf{Residual analysis}: Examine prediction errors to identify potential model limitations or data quality issues.
\end{itemize}

\noindent This comprehensive analysis enables us to develop predictive relationships that can forecast massive activation emergence patterns based solely on architectural specifications, providing insights into how design choices influence these critical phenomena during training.

\section*{Data Availability}

The complete processed dataset, including all fitted parameters, statistics, and figures, is publicly released under the MIT License at \url{https://huggingface.co/datasets/Aimpoint-Digital/pythia-massive-activations}.

In addition to our processed dataset, the following publicly available datasets and models were used to generate the results in this study:

\begin{itemize}
    \item \textbf{Pythia Scaling Suite} – An EleutherAI collection, available at: \url{https://huggingface.co/collections/EleutherAI/pythia-scaling-suite-64fb5dfa8c21ebb3db7ad2e1}
    \item \textbf{Red Pajamas} – Available at: \url{https://huggingface.co/datasets/togethercomputer/RedPajama-Data-1T-Sample}
\end{itemize}

\section*{Ethical Approval}
This study was reviewed by the Institutional Review Board of New York University and was determined to be exempt from further review (IRB protocol number: IRB-FY2025-10500).

\section*{Funding}
This work was not supported by any specific funding.

\bibliography{sample}

@inproceedings{Sun:massiveactivations:2024,
  author = {Sun, Mingjie and Chen, Xinlei and Kolter, J. Zico and Liu, Zhuang},
  year = {2024},
  title = {Massive Activations in Large Language Models},
  booktitle = {First Conference on Language Modeling},
  note = {arXiv preprint: \url{https://arxiv.org/abs/2402.17762}}
}

@misc{owen2025refinedanalysismassiveactivations,
      title={A Refined Analysis of Massive Activations in LLMs}, 
      author={Louis Owen and Nilabhra Roy Chowdhury and Abhay Kumar and Fabian Güra},
      year={2025},
      eprint={2503.22329},
      archivePrefix={arXiv},
      primaryClass={cs.CL},
      url={https://arxiv.org/abs/2503.22329}, 
}

@misc{oh2025housecardsmassiveweights,
      title={House of Cards: Massive Weights in LLMs}, 
      author={Jaehoon Oh and Seungjun Shin and Dokwan Oh},
      year={2025},
      eprint={2410.01866},
      archivePrefix={arXiv},
      primaryClass={cs.LG},
      url={https://arxiv.org/abs/2410.01866}, 
}

@misc{biderman2023pythiasuiteanalyzinglarge,
      title={Pythia: A Suite for Analyzing Large Language Models Across Training and Scaling}, 
      author={Stella Biderman and Hailey Schoelkopf and Quentin Anthony and Herbie Bradley and Kyle O'Brien and Eric Hallahan and Mohammad Aflah Khan and Shivanshu Purohit and USVSN Sai Prashanth and Edward Raff and Aviya Skowron and Lintang Sutawika and Oskar van der Wal},
      year={2023},
      eprint={2304.01373},
      archivePrefix={arXiv},
      primaryClass={cs.CL},
      url={https://arxiv.org/abs/2304.01373}, 
}

@misc{yu2025superweightlargelanguage,
      title={The Super Weight in Large Language Models}, 
      author={Mengxia Yu and De Wang and Qi Shan and Colorado J Reed and Alvin Wan},
      year={2025},
      eprint={2411.07191},
      archivePrefix={arXiv},
      primaryClass={cs.CL},
      url={https://arxiv.org/abs/2411.07191}, 
}

@misc{together2023redpajama_sample,
  author = {Together Computer},
  title  = {RedPajama: An Open-Source Recipe to Reproduce the LLaMA Training Dataset
            (RedPajama-Data-1T-Sample)},
  month  = {April},
  year   = {2023},
  url    = {https://github.com/togethercomputer/RedPajama-Data}
}

@misc{togethercomputer_redpajama_1Tsample_2023,
  author    = {Together Computer},
  title     = {{RedPajama-Data-1T-Sample}: a 1 B-token open corpus for LLM pre-training},
  year      = {2023},
  publisher = {Hugging Face},
  version   = {v1.0.0},
  url       = {https://huggingface.co/datasets/togethercomputer/RedPajama-Data-1T-Sample},
  note      = {Accessed 28 Jul 2025}
}

@article{akaike1974aic,
  author  = {Akaike, Hirotugu},
  title   = {A New Look at the Statistical Model Identification},
  journal = {IEEE Transactions on Automatic Control},
  year    = {1974},
  volume  = {19},
  number  = {6},
  pages   = {716--723},
  doi     = {10.1109/TAC.1974.1100705}
}

@misc{touvron2023llama2openfoundation,
      title={Llama 2: Open Foundation and Fine-Tuned Chat Models}, 
      author={Hugo Touvron and Louis Martin and Kevin Stone and Peter Albert and Amjad Almahairi and Yasmine Babaei and Nikolay Bashlykov and Soumya Batra and Prajjwal Bhargava and Shruti Bhosale and Dan Bikel and Lukas Blecher and Cristian Canton Ferrer and Moya Chen and Guillem Cucurull and David Esiobu and Jude Fernandes and Jeremy Fu and Wenyin Fu and Brian Fuller and Cynthia Gao and Vedanuj Goswami and Naman Goyal and Anthony Hartshorn and Saghar Hosseini and Rui Hou and Hakan Inan and Marcin Kardas and Viktor Kerkez and Madian Khabsa and Isabel Kloumann and Artem Korenev and Punit Singh Koura and Marie-Anne Lachaux and Thibaut Lavril and Jenya Lee and Diana Liskovich and Yinghai Lu and Yuning Mao and Xavier Martinet and Todor Mihaylov and Pushkar Mishra and Igor Molybog and Yixin Nie and Andrew Poulton and Jeremy Reizenstein and Rashi Rungta and Kalyan Saladi and Alan Schelten and Ruan Silva and Eric Michael Smith and Ranjan Subramanian and Xiaoqing Ellen Tan and Binh Tang and Ross Taylor and Adina Williams and Jian Xiang Kuan and Puxin Xu and Zheng Yan and Iliyan Zarov and Yuchen Zhang and Angela Fan and Melanie Kambadur and Sharan Narang and Aurelien Rodriguez and Robert Stojnic and Sergey Edunov and Thomas Scialom},
      year={2023},
      eprint={2307.09288},
      archivePrefix={arXiv},
      primaryClass={cs.CL},
      url={https://arxiv.org/abs/2307.09288}, 
}

@ARTICLE{2020SciPy-NMeth,
  author  = {Virtanen, Pauli and Gommers, Ralf and Oliphant, Travis E. and
            Haberland, Matt and Reddy, Tyler and Cournapeau, David and
            Burovski, Evgeni and Peterson, Pearu and Weckesser, Warren and
            Bright, Jonathan and {van der Walt}, St{\'e}fan J. and
            Brett, Matthew and Wilson, Joshua and Millman, K. Jarrod and
            Mayorov, Nikolay and Nelson, Andrew R. J. and Jones, Eric and
            Kern, Robert and Larson, Eric and Carey, C J and
            Polat, {\.I}lhan and Feng, Yu and Moore, Eric W. and
            {VanderPlas}, Jake and Laxalde, Denis and Perktold, Josef and
            Cimrman, Robert and Henriksen, Ian and Quintero, E. A. and
            Harris, Charles R. and Archibald, Anne M. and
            Ribeiro, Ant{\^o}nio H. and Pedregosa, Fabian and
            {van Mulbregt}, Paul and {SciPy 1.0 Contributors}},
  title   = {{{SciPy} 1.0: Fundamental Algorithms for Scientific
            Computing in Python}},
  journal = {Nature Methods},
  year    = {2020},
  volume  = {17},
  pages   = {261--272},
  adsurl  = {https://rdcu.be/b08Wh},
  doi     = {10.1038/s41592-019-0686-2},
}

@article{ColemanLi1996TRF,
  author       = {Coleman, T. F. and Li, Y.},
  title        = {An Interior, Trust Region Approach for Nonlinear Minimization Subject to Bounds},
  journal      = {SIAM Journal on Optimization},
  year         = {1996},
  volume       = {6},
  number       = {2},
  pages        = {418--445},
  doi          = {10.1137/0806023}
}

@inproceedings{bondarenko2023quantizable,
  title={Quantizable Transformers: Removing Outliers by Helping Attention Heads Do Nothing},
  author={Bondarenko, Yelysei and Nagel, Markus and Blankevoort, Tijmen},
  booktitle={NeurIPS},
  year={2023},
  url={https://arxiv.org/abs/2306.12929}
}

@article{he2024understanding,
  title={Understanding and minimising outlier features in transformer training},
  author={He, Bobby and Noci, Lorenzo and Paliotta, Daniele and Schlag, Imanol and Hofmann, Thomas},
  journal={Advances in Neural Information Processing Systems},
  volume={37},
  pages={83786--83846},
  year={2024}
}

@misc{ma2024activationsmattertrainingfreemethods,
      title={First Activations Matter: Training-Free Methods for Dynamic Activation in Large Language Models}, 
      author={Chi Ma and Mincong Huang and Ying Zhang and Chao Wang and Yujie Wang and Lei Yu and Chuan Liu and Wei Lin},
      year={2024},
      eprint={2408.11393},
      archivePrefix={arXiv},
      primaryClass={cs.CL},
      url={https://arxiv.org/abs/2408.11393}, 
}

@inproceedings{szatkowski_exploiting_2024,
	title = {Exploiting {Activation} {Sparsity} with {Dense} to {Dynamic}-k {Mixture}-of-{Experts} {Conversion}},
	volume = {37},
	url = {https://proceedings.neurips.cc/paper_files/paper/2024/file/4c2092ec0b1370cce3fb5965ab255fae-Paper-Conference.pdf},
	booktitle = {Advances in {Neural} {Information} {Processing} {Systems}},
	publisher = {Curran Associates, Inc.},
	author = {Szatkowski, Filip and Wójcik, Bartosz and Piórczyński, Mikoł aj and Scardapane, Simone},
	editor = {Globerson, A. and Mackey, L. and Belgrave, D. and Fan, A. and Paquet, U. and Tomczak, J. and Zhang, C.},
	year = {2024},
	pages = {43245--43273},
}

@article{lin2024duquant,
  title={Duquant: Distributing outliers via dual transformation makes stronger quantized llms},
  author={Lin, Haokun and Xu, Haobo and Wu, Yichen and Cui, Jingzhi and Zhang, Yingtao and Mou, Linzhan and Song, Linqi and Sun, Zhenan and Wei, Ying},
  journal={Advances in Neural Information Processing Systems},
  volume={37},
  pages={87766--87800},
  year={2024}
}

@inproceedings{kim2025peri,
  title={Peri-ln: Revisiting normalization layer in the transformer architecture},
  author={Kim, Jeonghoon and Lee, Byeongchan and Park, Cheonbok and Oh, Yeontaek and Kim, Beomjun and Yoo, Taehwan and Shin, Seongjin and Han, Dongyoon and Shin, Jinwoo and Yoo, Kang Min},
  booktitle={Forty-second International Conference on Machine Learning},
  year={2025}
}

@article{darcet2023vision,
  title={Vision transformers need registers},
  author={Darcet, Timoth{\'e}e and Oquab, Maxime and Mairal, Julien and Bojanowski, Piotr},
  journal={arXiv preprint arXiv:2309.16588},
  year={2023}
}

@inproceedings{
an2025systematic,
title={Systematic Outliers in Large Language Models},
author={Yongqi An and Xu Zhao and Tao Yu and Ming Tang and Jinqiao Wang},
booktitle={The Thirteenth International Conference on Learning Representations},
year={2025},
url={https://openreview.net/forum?id=rLX7Vyyzus}
}

@article{narang2024deepseek_v3,
  title={DeepSeek‑V3 Technical Report},
  author={DeepSeek Team},
  journal={arXiv preprint arXiv:2412.19437},
  year={2024},
  note={Details FP8 training for a 671B MoE model, using tile-wise scaling to handle activation outliers and maintain stability},
  url={https://arxiv.org/abs/2412.19437}
}

@article{dettmers2022llmint8,
  title={GPTQ and LLM.int8(): 1‑bit Quantization for Large Language Models},
  author={Dettmers, Tim and Thoppilan, Rajat and others},
  journal={NeurIPS},
  year={2022},
  note={Isolates activation outliers into high-precision for 8-bit inference},
  url={https://arxiv.org/abs/2202.00107}
}

@article{nrusimha2024activation_regularization,
  title={Mitigating the Impact of Outlier Channels for Language Model Quantization with Activation Regularization},
  author={Nrusimha, Aniruddha and Mishra, Mayank and Wang, Naigang and Alistarh, Dan and Panda, Rameswar and Kim, Yoon},
  journal={arXiv preprint arXiv:2404.03605},
  year={2024},
  note={Applies kurtosis-based regularization to mitigate activation outliers for W4A4 quantization},
  url={https://arxiv.org/abs/2404.03605}
}

@article{breiman2001random,
  title={Random forests},
  author={Breiman, Leo},
  journal={Machine learning},
  volume={45},
  number={1},
  pages={5--32},
  year={2001},
  publisher={Springer}
}

@inproceedings{chen2016xgboost,
  title={Xgboost: A scalable tree boosting system},
  author={Chen, Tianqi and Guestrin, Carlos},
  booktitle={Proceedings of the 22nd acm sigkdd international conference on knowledge discovery and data mining},
  pages={785--794},
  year={2016}
}

@article{lundberg2017unified,
  title={A unified approach to interpreting model predictions},
  author={Lundberg, Scott M and Lee, Su-In},
  journal={Proceedings of the 31st International Conference on Neural Information Processing Systems},
  pages = {4768–4777},
  year={2017}
}

@article{friedman2001greedy,
  title={Greedy function approximation: a gradient boosting machine},
  author={Friedman, Jerome H},
  journal={Annals of statistics},
  pages={1189--1232},
  year={2001},
}

@article{vaswani2017attention,
  title={Attention is all you need},
  author={Vaswani, Ashish and Shazeer, Noam and Parmar, Niki and Uszkoreit, Jakob and Jones, Llion and Gomez, Aidan N and Kaiser, {\L}ukasz and Polosukhin, Illia},
  journal={Advances in neural information processing systems},
  pages = {6000–6010},
  year={2017}
}

@article{devlin2018bert,
  title={Bert: Pre-training of deep bidirectional transformers for language understanding},
  author={Devlin, Jacob and Chang, Ming-Wei and Lee, Kenton and Toutanova, Kristina},
  journal={arXiv preprint arXiv:1810.04805},
  year={2018}
}

@article{radford2018improving,
  title={Improving language understanding by generative pre-training},
  author={Radford and Narasimhan and Salimans, Sutskever},
  journal={OpenAI Technical Report},
  year={2018}
}

@article{power2022grokking,
  title={Grokking: Generalization beyond overfitting on small algorithmic datasets},
  author={Power, Alethea and Burda, Yura and Edwards, Harri and Babuschkin, Igor and Misra, Vedant},
  journal={arXiv preprint arXiv:2201.02177},
  year={2022}
}

@article{liu2022towards,
  title={Towards understanding grokking: An effective theory of representation learning},
  author={Liu, Ziming and Kitouni, Ouail and Nolte, Niklas S and Michaud, Eric and Tegmark, Max and Williams, Mike},
  journal={Advances in Neural Information Processing Systems},
  pages={34651--34663},
  year={2022}
}

@inproceedings{liu2021swin,
  title={Swin transformer: Hierarchical vision transformer using shifted windows},
  author={Liu, Ze and Lin, Yutong and Cao, Yue and Hu, Han and Wei, Yixuan and Zhang, Zheng and Lin, Stephen and Guo, Baining},
  booktitle={IEEE/CVF International Conference on Computer Vision (ICCV)},
  pages={9992--10002},
  year={2021}
}

@article{choromanski2020rethinking,
  title={Rethinking attention with performers},
  author={Choromanski, Krzysztof and Likhosherstov, Valerii and Dohan, David and Song, Xingyou and Gane, Andreea and Sarlos, Tamas and Hawkins, Peter and Davis, Jared and Mohiuddin, Afroz and Kaiser, Lukasz and others},
  journal={arXiv preprint arXiv:2009.14794},
  year={2020}
}

@misc{yue2024wkvquant,
      title={WKVQuant: Quantizing Weight and Key/Value Cache for Large Language Models}, 
      author={Yuxuan Yue and Zhihang Yuan and Haojie Duanmu and Sifan Zhou and Jianlong Wu and Liqiang Nie},
      year={2024},
      eprint={2402.12065},
      archivePrefix={arXiv},
      primaryClass={cs.LG},
      url={https://arxiv.org/abs/2402.12065},
}

@article{ba2016layer,
  title={Layer normalization},
  author={Ba, Jimmy Lei and Kiros, Jamie Ryan and Hinton, Geoffrey E},
  journal={arXiv preprint arXiv:1607.06450},
  year={2016}
}

@article{su2024roformer,
  title={RoFormer: Enhanced transformer with rotary position embedding},
  author={Su, Jianlin and Lu, Yu and Pan, Shanfeng and Wen, Bo and Liu, Yunfeng, and Murtadha, Adhmed},
  journal={Neurocomputing},
  volume={568},
  pages={127063},
  year={2024}
}

@article{press2021train,
  title={Train short, test long: Attention with linear biases enables input length extrapolation},
  author={Press, Ofir and Smith, Noah A and Lewis, Mike},
  journal={arXiv preprint arXiv:2108.12409},
  year={2021}
}

@article{enigmallm,
  title={From Attention to Activation: Unravelling the Enigmas of Large Language Models},
  author={Prannay Kaul and Chengcheng Ma and Ismail Elezi and Jiankang Deng},
  journal={arXiv preprint arXiv:2410.17174},
  year={2024}
}

@misc{jin2025massivevalues,
      title={Massive Values in Self-Attention Modules are the Key to Contextual Knowledge Understanding}, 
      author={Mingyu Jin and Kai Mei and Wujiang Xu and Mingjie Sun and Ruixiang Tang and Mengnan Du and Zirui Liu and Yongfeng Zhang},
      year={2025},
      eprint={2502.01563},
      archivePrefix={arXiv},
      primaryClass={cs.CL},
      url={https://arxiv.org/abs/2502.01563},
}

@misc{gan2025unleashing,
      title={Unleashing Diffusion Transformers for Visual Correspondence by Modulating Massive Activations}, 
      author={Chaofan Gan and Yuanpeng Tu and Xi Chen and Tieyuan Chen and Yuxi Li and Mehrtash Harandi and Weiyao Lin},
      year={2025},
      eprint={2505.18584},
      archivePrefix={arXiv},
      primaryClass={cs.CV},
      url={https://arxiv.org/abs/2505.18584},
      note={Under Review}
}

@misc{zuhri2025softpick,
      title={Softpick: No Attention Sink, No Massive Activations with Rectified Softmax}, 
      author={Zayd M. K. Zuhri and Erland Hilman Fuadi and Alham Fikri Aji},
      year={2025},
      eprint={2504.20966},
      archivePrefix={arXiv},
      primaryClass={cs.LG},
      url={https://arxiv.org/abs/2504.20966},
}

@misc{zhao2025activation,
      title={Activation Control for Efficiently Eliciting Long Chain-of-thought Ability of Language Models}, 
      author={Zekai Zhao and Qi Liu and Kun Zhou and Zihan Liu and Yifei Shao and Zhiting Hu and Biwei Huang},
      year={2025},
      eprint={2505.17697},
      archivePrefix={arXiv},
      primaryClass={cs.CL},
      url={https://arxiv.org/abs/2505.17697},
      note={Under review}
}

@misc{xu2024tracking,
      title={Tracking the Feature Dynamics in LLM Training: A Mechanistic Study}, 
      author={Yang Xu and Hengguan Huang and Yi Wang and Hao Wang},
      year={2024},
      eprint={2412.17626},
      archivePrefix={arXiv},
      primaryClass={cs.LG},
      url={https://arxiv.org/abs/2412.17626},
}

@misc{yang2024mitigating,
      title={Mitigating Quantization Errors Due to Activation Spikes in GLU-Based LLMs}, 
      author={Jaewoo Yang and Hayun Kim and Younghoon Kim},
      year={2024},
      eprint={2405.14428},
      archivePrefix={arXiv},
      primaryClass={cs.CL},
      url={https://arxiv.org/abs/2405.14428},
}

@misc{zhang2019rmsnorm,
      title={Root Mean Square Layer Normalization}, 
      author={Biao Zhang and Rico Sennrich},
      year={2019},
      eprint={1910.07467},
      archivePrefix={arXiv},
      primaryClass={cs.CL},
      url={https://arxiv.org/abs/1910.07467},
}

@misc{zhu2025transformers,
      title={Transformers without Normalization}, 
      author={Jiachen Zhu and Xinlei Chen and Kaiming He and Yann LeCun and Zhuang Liu},
      year={2025},
      eprint={2503.10622},
      archivePrefix={arXiv},
      primaryClass={cs.LG},
      url={https://arxiv.org/abs/2503.10622},
      note={CVPR 2025}
}

@article{brown2020language,
  title={Language models are few-shot learners},
  author={Brown, Tom and Mann, Benjamin and Ryder, Nick and Subbiah, Melanie and Kaplan, Jared D and Dhariwal, Prafulla and Neelakantan, Arvind and Shyam, Pranav and Sastry, Girish and Askell, Amanda and others},
  journal={Advances in neural information processing systems},
  volume={33},
  pages={1877--1901},
  year={2020}
}

@book{zhang2023dive,
  title={Dive into deep learning},
  author={Zhang, Aston and Lipton, Zachary C and Li, Mu and Smola, Alexander J},
  year={2023},
  publisher={Cambridge University Press}
}

@article{yeo2000new,
  title={A new family of power transformations to improve normality or symmetry},
  author={Yeo, In-Kwon and Johnson, Richard A},
  journal={Biometrika},
  volume={87},
  number={4},
  pages={954--959},
  year={2000},
  publisher={Oxford University Press}
}

\end{document}